\documentclass[11pt]{article}

\usepackage[preprint]{acl}

\usepackage{times}
\usepackage{latexsym}

\usepackage[T1]{fontenc}

\usepackage[utf8]{inputenc}

\usepackage{microtype}

\usepackage{inconsolata}

\usepackage{graphicx}
\usepackage{amsmath}
\usepackage{array}
\usepackage{tabularx}
\usepackage{booktabs}
\title{Sense Representations Are Inducible Interfaces}

\author{Jan Christian Blaise Cruz\textsuperscript{1,2} \textnormal{and} Alham Fikri Aji\textsuperscript{1,2}\\
  MBZUAI\textsuperscript{1} \quad SEACrowd\textsuperscript{2} \\
  \texttt{\{jan.cruz,alham.fikri\}@mbzuai.ac.ae}}

\begin{document}
\maketitle

\begin{abstract}
Sense representations (explicit, per-token meaning decompositions) are useful for disambiguation, steering, and cross-lingual alignment, but existing approaches require models to be pretrained with sense structure baked in. We introduce ACROS, which \emph{induces} an explicit sense pathway into a frozen pretrained decoder LM through a gated residual addition. On SmolLM2-360M, ACROS preserves base LM quality while supporting three uses of the same induced variables: zero-shot word-sense disambiguation (64.95 F1 on Raganato ALL, competitive with the WordNet first-sense heuristic), low-KL lexical steering across 5,161 CoInCo cases where a simple non-oracle proxy recovers about 90\% of positive shifts, and SENSIA cross-lingual adaptation to four languages (mean R@1 0.988, target FLORES PPL 7.94). ACROS makes sense representations an inducible interface for ordinary pretrained LMs.
\end{abstract}

\section{Introduction}
\label{sec:intro}

Words carry multiple meanings, and a model that can explicitly represent, manipulate, and align those meanings gains capabilities that dense, opaque hidden states do not easily support. Sense-aware systems already exist for individual tasks: disambiguation \citep{blevins-zettlemoyer-2020-moving,loureiro-etal-2022-lmms}, representation steering \citep{turner-etal-2023-activation,rimsky-etal-2024-steering}, and cross-lingual alignment \citep{cruz-etal-2026-multilinguality}. But each is tied to its own architecture or training procedure. No single system exposes the same sense objects for measurement, intervention, and alignment.

Pretrained LMs are the natural unifying framework, as they already encode rich lexical structure \citep{petroni-etal-2019-language}, but keep it locked inside dense hidden states. Backpack Language Models \citep{hewitt-etal-2023-backpack} come closest to exposing it, giving each token multiple explicit sense vectors. The catch is that Backpacks must be pretrained as Backpacks. Most strong checkpoints are ordinary decoders, and as we show in Section~\ref{sec:background}, converting one into a Backpack creates a representational bottleneck that destroys prediction quality.

\begin{figure}[t]
  \centering
  \includegraphics[width=\columnwidth]{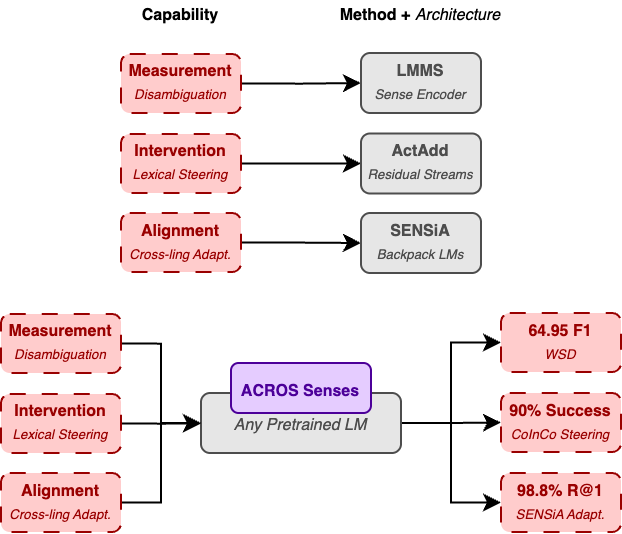}
  \caption{\textbf{ACROS unifies measurement, intervention, and alignment in one residual sense interface designed for pretrained decoder LMs.} Existing sense-based capabilities were previously tied to different architectures, with no single interface that supports all three. ACROS inserts a residual sense pathway so the same induced variables can be read, steered, and aligned.}
  \label{fig:thesis}
\end{figure}

In this paper, we propose \textbf{\underline{A}r\underline{c}hitecture-agnostic \underline{R}esidual Induction \underline{o}f \underline{S}ense Representations} (ACROS), a method that induces an explicit sense pathway into a frozen decoder LM through a gated residual addition, without modifying the original prediction path. Our contributions are:

\begin{enumerate}
  \item A residual sense-induction method that adds a reusable sense interface to a pretrained decoder LM while preserving its prediction quality.
  \item A unified evaluation showing that the \emph{same} induced variables support word-sense disambiguation, low-KL lexical steering, and SENSIA cross-lingual adaptation: three roles that previously required separate systems.
  \item Ablations examining architecture portability across three decoder families, the role of the explicit sense pathway relative to dense hidden-state readout, and interface capacity and scale.
\end{enumerate}

\section{Background and Motivation}
\label{sec:background}

Sense representations serve three roles in NLP, each supported by separate systems. \textbf{Measurement} systems disambiguate words in context through gloss-informed embeddings \citep{blevins-zettlemoyer-2020-moving,loureiro-etal-2022-lmms} or multi-sense distillation \citep{wang-etal-2025-multi-sense,wang-etal-2026-dskd}. \textbf{Intervention} methods steer model behavior by modifying internal representations \citep{keskar-etal-2019-ctrl,dathathri-etal-2020-plug,yang-klein-2021-fudge,turner-etal-2023-activation,rimsky-etal-2024-steering,zou-etal-2023-representation,arad-etal-2025-saes,chalnev-etal-2024-improving}, though recent evaluations find this can be brittle \citep{pmlr-v267-wu25a,silva-etal-2025-steering}. \textbf{Alignment} methods use sense structure to anchor cross-lingual adaptation \citep{cruz-etal-2026-multilinguality,hewitt-etal-2023-backpack}. However, a WSD model cannot steer generation; a steering method cannot be evaluated as WSD; neither provides sense anchors for alignment. For senses to become a \emph{general-purpose interface}, all three roles must be supported by the same representations inside the same model.

\subsection{The Backpack Bottleneck}

Backpack Language Models \citep{hewitt-etal-2023-backpack} come closest to this goal. Each token has $K$ explicit sense vectors with learned contextualization weights, and the same machinery can be measured, edited \citep{hewitt-etal-2024-model-editing}, and used for adaptation. However, Backpacks must be pretrained in that form. Most strong LM checkpoints are ordinary decoders, so the practical question is whether an existing LM can be \emph{converted} into a Backpack.

We test this directly by converting SmolLM2-360M \citep{allal-etal-2025-smollm2}\footnote{SmolLM2 was chosen because it is text-only (which eliminates potential visually-learned confounds), English-only (which eliminates multilingual confounds relevant to later evaluation via SENSIA), and has a fully-open pretraining corpus (which prevents test-set leakage concerns).}, an English-only Llama-style decoder, into converted Backpack models through continued training, distillation, and frozen-backbone controls. Table~\ref{tab:conversion-main} tells a consistent story: every conversion variant learns non-collapsed sense vectors, but sharply degrades ordinary LM behavior. FLORES perplexity \citep{goyal-etal-2022-flores} jumps from 25.1 to between 196 and 357; LAMA factual recall \citep{petroni-etal-2019-language} falls well below half of the original. The frozen-backbone control is especially telling: even when the original backbone and LM head are kept fixed and only the sense-side modules are trained, prediction quality still collapses. The failure is not in the training procedure; it is in the output architecture. Appendix~\ref{app:conversion} gives extended diagnostics across additional conversion variants.

\begin{table}[!t]
  \centering
  \small
  \begin{tabularx}{\columnwidth}{@{}Xrrr@{}}
    \toprule
    Variant & PPL & LAMA@1 & Sense Sep. \\
    \midrule
    Original              & 25.1  & 0.3155 & --      \\
    BP + CPT              & 356.9 & 0.0695 & -0.0833 \\
    BP + Distil.          & 196.3 & 0.0420 & -0.1275 \\
    BP + Distil. + Freeze & 236.9 & 0.1245 & -0.0308 \\
    \bottomrule
  \end{tabularx}
  \caption{\textbf{Converting a pretrained LM to a Backpack model creates sense representations, but the resulting model loses the original LM's predictive ability.} PPL is English FLORES perplexity; LAMA@1 is filtered factual cloze accuracy. Sense separation (Sense Sep.) is the mean off-diagonal cosine between sense vectors, where lower or negative values indicate that slots are not collapsing onto the same direction.}
  \label{tab:conversion-main}
\end{table}

\begin{figure}[t]
  \centering
  \includegraphics[width=\columnwidth]{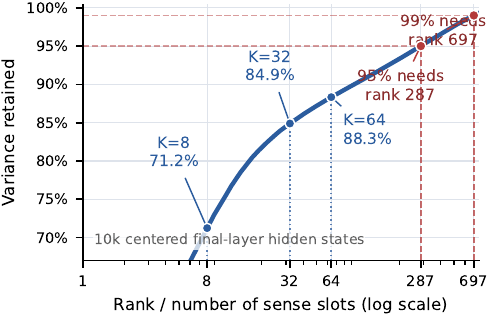}
  \caption{\textbf{Conversion fails because practical sense counts cannot span the hidden-state space the LM already uses.} On SmolLM2-360M final-layer states, practical sense counts ($K{=}8$, 32, 64) retain far less variance than the ranks needed for 95\% or 99\% coverage, explaining why Backpack conversion harms prediction quality.}
  \label{fig:svd}
\end{figure}

\paragraph{Why conversion fails.} The bottleneck has a geometric explanation. In a converted Backpack, the pre-logit representation is a convex mixture of $K$ sense vectors: $h_q^{\mathrm{BP}}=\sum_k \alpha_{q,k}v_{q,k}$, with $\alpha_{q,k}\ge 0$ and $\sum_k \alpha_{q,k}=1$. This forces all prediction-relevant information through a space spanned by at most $K$ vectors. Figure~\ref{fig:svd} shows the problem concretely: we collect 10,000 final-layer hidden states from the pretrained SmolLM2 backbone on its own pretraining corpus, FineWeb, and measure how much variance the top $k$ singular components capture. At the canonical sense count ($K{=}32$), only 84.9\% of the variance is retained; reaching 95\% requires rank 287, and 99\% requires rank 697. A converted Backpack with practical $K$ simply cannot represent the full space that the pretrained LM uses for prediction.

This is not a formal impossibility result (Backpack sense vectors are input-dependent and could in principle span a larger space), but it explains why conversion consistently fails even under favorable conditions. The residual design of ACROS avoids this bottleneck entirely: instead of replacing the hidden state with a sense mixture, we \emph{add} a sense mixture to the existing hidden state.

\section{ACROS}
\label{sec:method}

ACROS adds an explicit sense pathway to a pretrained decoder LM while leaving the original prediction path intact. The design principle is simple: rather than forcing all prediction through a sense mixture (which creates the bottleneck shown in Section~\ref{sec:background}), we add a sense mixture \emph{alongside} the frozen base hidden state and let a learned gate control how much influence it has.

\begin{figure}[t]
  \centering
  \includegraphics[width=\columnwidth]{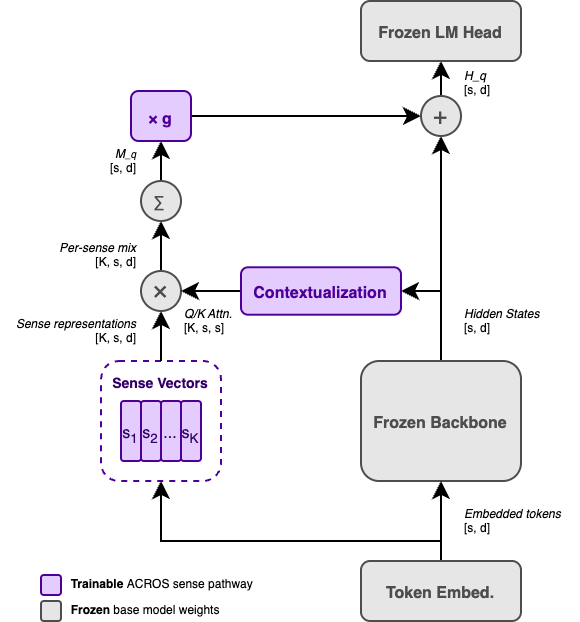}
  \caption{\textbf{ACROS keeps the original LM intact and learns a gated residual sense path beside it.} Token embeddings feed both the frozen backbone (producing base hidden state $B_q$) and a trainable sense MLP (producing $K$ sense vectors per token). A contextualization layer computes per-sense Q/K attention weights from the backbone's hidden states, which are multiplied with the sense vectors and summed to form the sense mixture $M_q$. A scalar gate $g$, initialized to zero, controls the residual addition $H_q = B_q + g \cdot M_q$ before the frozen LM head. Gray = frozen; purple = trainable.}
  \label{fig:architecture}
\end{figure}

\paragraph{Architecture.} The sense MLP maps token embeddings to $K$ source-token sense vectors\footnote{We use \textit{sense} operationally to mean a continuous residual direction that captures a distinct lexical aspect of a token. These vectors are induced from the language-modeling objective and are not discrete lexicographic entries; the WSD evaluations and qualitative neighbor analyses show that they nevertheless align with human sense distinctions.} $E_{k,j}$. A contextualization layer maps frozen backbone states to per-sense query/key attention weights $C_{k,q,j}$, selecting which source-token senses enter each position's residual update. For base hidden state $B_q$, scalar gate $g$, and output matrix $W$:
\begin{align}
  M_q &= \sum_k \sum_j C_{k,q,j} E_{k,j}, \\
  H_q &= B_q + g M_q, \\
  \mathrm{logits}_q &= H_q W^\top.
\end{align}
The gate is initialized to zero, so ACROS initially matches the base LM exactly and learns only whether to open the residual sense path. A diagram of the architecture is given in Figure~\ref{fig:architecture}. Full implementation details, including the contextualization layer parameterization, parameter counts, and latency measurements, are given in Appendix~\ref{app:architecture-details}.

\paragraph{Induction.}
The induced model starts from the same checkpoint, freezes the original backbone and LM head, and trains only the sense MLP, contextualization layer, and scalar gate against the frozen base LM teacher.

For sequence labels $y_t$, binary token mask $m_t$, teacher logits $z_t^T$, ACROS logits $z_t^S$, and normalization constant $M=\sum_t m_t$, the full induction loss is
\begin{equation}
  \mathcal{L}
  = \alpha \mathcal{L}_{\mathrm{CLM}}
  + (1-\alpha)\tau^2\mathcal{L}_{\mathrm{KD}}
  + \lambda_{\mathrm{div}}\mathcal{L}_{\mathrm{div}} .
\end{equation}
The knowledge-distillation term keeps ACROS close to the pretrained decoder's output distribution. With $p^T_{t,\tau}=\mathrm{softmax}(z_t^T/\tau)$ and $p^S_{t,\tau}=\mathrm{softmax}(z_t^S/\tau)$:
\begin{equation}
  \mathcal{L}_{\mathrm{KD}}
  = \frac{1}{M}\sum_t m_t
    \mathrm{KL}\!\left(p^T_{t,\tau}\Vert p^S_{t,\tau}\right).
\end{equation}
The causal language modeling (CLM) term keeps the induced model grounded in ordinary next-token prediction:
\begin{equation}
  \mathcal{L}_{\mathrm{CLM}}
  = -\frac{1}{M}\sum_t m_t
    \log \mathrm{softmax}(z_t^S)_{y_t}.
\end{equation}
The diversity term discourages the $K$ same-token sense vectors from collapsing into one shared direction:
\begin{equation}
  \mathcal{L}_{\mathrm{div}}
  = \frac{1}{M}\sum_t m_t
    \frac{1}{K(K-1)}
    \sum_{k\ne \ell}
    \left(\hat E_{k,t}^{\top}\hat E_{\ell,t}\right)^2 ,
\end{equation}
where $\hat E_{k,t}=E_{k,t}/\lVert E_{k,t}\rVert_2$. We fix $\lambda_{\mathrm{div}}=0.005$ throughout; ablating this coefficient is left to future work. Since the base path is frozen and the gate starts at zero, induction only learns a residual sense decomposition that can help the existing prediction path.

\begin{table}[!t]
  \centering
  \small
  \setlength{\tabcolsep}{0pt}
  \begin{tabular*}{\columnwidth}{@{\extracolsep{\fill}}lccc@{}}
    \toprule
    Metric & Original & Frozen BP & ACROS \\
    \midrule
    FLORES PPL $\downarrow$ & 25.1 & 236.9 & \textbf{25.09} \\
    LAMA@1 $\uparrow$ & 0.3155 & 0.1245 & \textbf{0.3175} \\
    Sense separation & -- & -0.0308 & -0.0105 \\
    WS-353 max $\uparrow$ & -- & 0.139 & \textbf{0.487} \\
    \bottomrule
  \end{tabular*}
  \caption{\textbf{This residual path preserves LM quality while learning useful sense representations.} Unlike Frozen BP, a Backpack LM distilled with a frozen SmolLM backbone, ACROS preserves base-level FLORES PPL and LAMA@1 while improving lexical/sense diagnostics, showing that residual induction adds an interface without damaging the prediction path.}
  \label{tab:intrinsic}
\end{table}

\paragraph{Intrinsic Results.} We apply ACROS to SmolLM2-360M with 10.5B tokens of FineWeb \citep{allal-etal-2025-smollm2}. Table~\ref{tab:intrinsic} shows the payoff of the residual design. ACROS preserves the base model's perplexity and factual recall (LAMA@1 even slightly exceeds the original at 0.3175 vs.\ 0.3155) while exposing a sense interface with strong lexical content (WS-353 \citep{finkelstein-etal-2001-placing} Spearman correlation 0.487). The converted Backpack, by contrast, retains some sense structure but at the cost of a $9.5\times$ increase in perplexity.

\begin{table}[t]
  \centering
  {\small
  \setlength{\tabcolsep}{3pt}
  \renewcommand{\arraystretch}{1.1}
  \begin{tabularx}{\columnwidth}{@{}p{0.33\columnwidth}>{\raggedright\arraybackslash}X@{}}
    \toprule
    \textbf{Field} & \textbf{Example} \\
    \midrule
    \textbf{Target word} &
    \texttt{article}. \\
    \midrule
    \textbf{Task} &
    Disambiguate the word. \\
    \textbf{Sentence} &
    ``\ldots read the \texttt{article} by Caleb Nelson \ldots'' \\
    \textbf{Method} &
    Cosine sim. of target and gloss ACROS activations. \\
    \textbf{Matched gloss} &
    \texttt{article\%1:10:00::}: ``nonfictional prose forming an independent part of a publication.'' \\
    \midrule
    \textbf{Task} &
    Increase next-token odds of related substitutes. \\
    \textbf{Sentence} &
    ``I remembered an \texttt{article} in The Medusa \ldots'' \\
    \textbf{Method} &
    Boost sense vector 14 by $1.2\times$. \\
    \textbf{Mass change} &
    $\Delta = +4.36\mathrm{e}{-7}$ (essay, piece, \ldots). \\
    \midrule
    \textbf{Task} &
    Convert \texttt{eng} model to \texttt{ind}. \\
    \textbf{Method} &
    SENSIA. \\
    \textbf{Prompt} &
    ``Artikel itu menjelaskan bahwa'' \\
    \textbf{Generation} &
    ``\ldots pemerintah AS telah menyatakan bahwa mereka tidak akan menerima pembayaran dari pemerintah Suriah untuk senjata yang dibeli dari Rusia.''\\
    \bottomrule
  \end{tabularx}
  \renewcommand{\arraystretch}{1.0}}
  \caption{\textbf{The resulting induced sense interface supports measurement, intervention, and cross-lingual adaptation.} (Top) The active sense vector successfully matches words in context to the correct WSD gloss. (Middle) Interventions on the selected vector increase likelihood for related substitutes. (Bottom) The induced senses act as anchors for cross-lingual adaptation via SENSIA, generating coherent Indonesian text.}
  \label{tab:qual-trace}
\end{table}

\begin{table}[t]
  \centering
  {\small
  \setlength{\tabcolsep}{4pt}
  \renewcommand{\arraystretch}{1.05}
  \begin{tabularx}{\columnwidth}{@{}r >{\raggedright\arraybackslash}p{0.32\columnwidth} >{\raggedright\arraybackslash}X@{}}
    \toprule
    \textbf{Sense} & \textbf{Nearest words} & \textbf{Reading} \\
    \midrule
    14 & article, articles, published &
    Published written piece. ``Article'' as a written work. \\
    6  & news, newspaper, press &
    News article. ``Article'' in a journalistic context. \\
    21 & ruling, prosecute, judiciary &
    Legal/institutional. ``Article'' in regulatory or legal language. \\
    9  & topic, discussion, interests &
    Topic or matter. ``Article'' as a subject of discussion. \\
    25 & substance, content, contribution &
    Content or substance. ``Article'' by what it conveys. \\
    \bottomrule
  \end{tabularx}
  \renewcommand{\arraystretch}{1.0}}
  \caption{\textbf{Different sense slots represent different facets of \texttt{article}.} Each slot retrieves a distinct filtered single-token neighborhood by cosine to the source-token sense vector, showing that ACROS decomposes the word across written-work, journalistic, legal, topical, and content-oriented readings.}
  \label{tab:article-sense-neighbors}
\end{table}


\paragraph{Qualitative Trace.}
Table~\ref{tab:qual-trace} traces a single word --- \texttt{article} --- through all three roles that the induced interface supports. For measurement, cosine similarity between ACROS activation vectors matches the word in context to the correct WordNet gloss. For intervention, boosting sense vector 14 by $1.2\times$ shifts probability mass toward human-judged substitutes; Section~\ref{sec:coinco} tests this systematically over 5,161 cases. For alignment, the same sense interface anchors SENSIA adaptation, and the originally English-only model generates coherent Indonesian text; Section~\ref{sec:sensia} evaluates this across four languages. Table~\ref{tab:article-sense-neighbors} gives a complementary view: the top sense slots for \texttt{article} cluster with distinct lexical neighborhoods (written-work, journalistic, legal, topical, and content-oriented readings), showing that the $K$ slots capture different facets of the word.

\section{Experiments}
\label{sec:experiments}

We evaluate ACROS through the three roles that motivate it: can the induced sense pathway be read as a semantic measurement, intervened on as a local causal handle, and reused as an alignment substrate for cross-lingual adaptation?

\subsection{WSD: Senses as Semantic \underline{Measurements}}
\label{sec:wsd}

\paragraph{Research Question.} Do induced ACROS sense activations recover lexical sense distinctions without WordNet or SemCor supervision during induction?

\paragraph{Methodology.} We evaluate on Raganato ALL \citep{raganato-etal-2017-word}, using WordNet only at evaluation time. ACROS scores each candidate through its induced interface: target occurrences and \texttt{lemma: gloss} probes are represented by L2-normalized $K$-dimensional sense-activation vectors (Figure~\ref{fig:wsd-format}), and we predict
\[
  \hat{s}=\arg\max_s \cos(a_{\mathrm{ctx}}, a_{\mathrm{gloss}}(s)).
\]
All 7,253 instances are answered without backoff or coverage failures. The same-family control, Base SmolLM2, scores the same glosses by length-normalized conditional likelihood, separating ordinary gloss preference from induced sense matching.

\begin{figure}[t]
  \centering
  \fbox{\begin{minipage}{0.92\columnwidth}
    \small
    \raggedright
    \textbf{Target context:} $\ldots$ the \texttt{bank} approved the loan $\ldots$\\
    \textbf{Candidate gloss prompt:} \texttt{bank: a financial institution}\\
    \textbf{Score:} cosine between target activation and gloss activation
  \end{minipage}}
  \caption{\textbf{For measurement, ACROS matches contextual target activations to WordNet gloss activations.} Each candidate WordNet sense is encoded as a \texttt{lemma: gloss} prompt and compared against the contextual target activation in the same interface that later supports steering and alignment.}
  \label{fig:wsd-format}
\end{figure}

\paragraph{Results.} Table~\ref{tab:wsd} shows ACROS performing competitive zero-shot WSD without sense supervision. ACROS reaches 64.95 F1 without any WordNet or SemCor supervision during induction, competitive with the WordNet first-sense heuristic (65.2) and 8.71 points above the same-family gloss-likelihood control (McNemar $p=1.31\mathrm{e}{-38}$; bootstrap 95\% CI $[+7.39,+10.01]$). Supervised systems score higher as expected. Section~\ref{sec:ablations} compares ACROS senses against dense hidden states.

\begin{table}[t]
  \centering
  {\small
  \begin{tabular}{llr}
    \toprule
    Method & Supervision & F1 \\
    \midrule
    BEM & SemCor + WordNet & 79.0 \\
    LMMS2048 & SemCor & 75.4 \\
    WordNet S1 & WordNet heuristic & 65.2 \\
    ACROS & zero-shot & 64.95 \\
    Base SmolLM2 & zero-shot gloss LM & 56.24 \\
    \bottomrule
  \end{tabular}}
  \caption{\textbf{ACROS learns strong sense representations that achieve competitive zero-shot WSD performance without WordNet or SemCor supervision.} WordNet S1 is a knowledge-based heuristic; BEM \citep{blevins-zettlemoyer-2020-moving} and LMMS2048 \citep{loureiro-etal-2022-lmms} use sense-annotated resources and are included as supervised context. Base SmolLM2 is a same-family gloss-likelihood control.}
  \label{tab:wsd}
\end{table}

\subsection{CoInCo: Senses as \underline{Intervention} Handles}
\label{sec:coinco}

\paragraph{Research Question.} Can a selected ACROS sense contribution causally move a target word's substitute distribution while keeping the next-token distribution local?

\paragraph{Methodology.} CoInCo \citep{kremer-etal-2014-substitutes} provides lexical-substitution judgments for target words in context. After filtering for unambiguous anchors and one-token substitutes, we score 5,161 cases as next-token prompts (Figure~\ref{fig:coinco-format}). We measure the change in log weighted substitute mass, $m(z,Y)=\log\sum_{y\in Y} w_y\, p_z(y)$, where weights are proportional to annotator counts. Success means $\Delta = m(z_{\mathrm{int}},Y)-m(z_{\mathrm{base}},Y)>0$. We use this rather than GAP or P@1 because those evaluate candidate rankings, while our experiment measures a distributional shift.

\paragraph{Intervention.} The intervention is target-token-local. For target word at position $j$ and answer position $q$, we boost one sense contribution analytically:
\begin{align}
\Delta h_k &= g(b-1)\,C_{k,q,j}\,E_{k,j}, \notag\\
z_{\mathrm{int}} &= z_{\mathrm{base}} + \Delta h_k\, W^\top. \notag
\end{align}
We compare five selectors: target-best (oracle with gold substitute access), self top-$k$ (non-oracle proxy selecting from the model's own one-token top predictions, no gold access), and three non-target-aware controls (contribution-norm, random, norm). We report $D_{\mathrm{KL}}(p_{\mathrm{base}}\|p_{\mathrm{int}})$.
\begin{figure}[t]
  \centering
  \fbox{\begin{minipage}{0.92\columnwidth}
    \small
    \raggedright
    \textbf{Context:} The reporter filed the \texttt{article} before noon.\\
    \textbf{Prompt:} A context-appropriate substitute for \texttt{article} is\\
    \textbf{Human substitute multiset:} story $\times$3, piece $\times$2, report, item\\
    \textbf{Intervention:} boost one target-token sense at the answer position
  \end{minipage}}
  \caption{\textbf{For intervention, ACROS tests whether a sense edit raises human substitute mass while preserving the base distribution.} We score interventions by how much they increase frequency-weighted probability mass on human CoInCo substitutes while keeping KL to the original next-token distribution low.}
  \label{fig:coinco-format}
\end{figure}

\paragraph{Results.} Table~\ref{tab:coinco} shows that boosting the right ACROS sense increases human substitute mass. Target-best moves positive substitute mass in every case. The non-oracle self top-$k$ proxy recovers substantial signal ($\Delta{=}{+}0.001109$, 89.8\% success) without access to gold annotations, while contribution-norm, random, and norm controls remain near chance, confirming that semantic alignment, not intervention magnitude, determines selector quality. We compare against dense hidden-state controls in Section~\ref{sec:ablations}.

\begin{table}[t]
  \centering
  {\small
  \begin{tabular}{lrrr}
    \toprule
    Selector & Delta & Succ.\,(\%) & KL \\
    \midrule
    target-best$^\dagger$ & $+1.64\mathrm{e}{-3}$ & 100.0 & $5.69\mathrm{e}{-7}$ \\
    self top-$k$          & $+1.11\mathrm{e}{-3}$ & 89.8  & $7.12\mathrm{e}{-7}$ \\
    contrib.-norm         & $+5.00\mathrm{e}{-6}$ & 49.6  & $8.24\mathrm{e}{-7}$ \\
    random                & $-8.00\mathrm{e}{-6}$ & 50.6  & $1.08\mathrm{e}{-7}$ \\
    norm                  & $-3.70\mathrm{e}{-5}$ & 48.4  & $2.46\mathrm{e}{-7}$ \\
    \bottomrule
  \end{tabular}}
  \caption{\textbf{Sense interventions successfully steer CoInCo substitutes, with self top-$k$ recovering much of the oracle effect.} Delta is mean log substitute-mass change at boost 1.2 over 5,161 cases. $\dagger$Oracle with gold access.}
  \label{tab:coinco}
\end{table}

\subsection{SENSIA: Senses as Anchors for Cross-Lingual \underline{Alignment}}
\label{sec:sensia}

\paragraph{Research Question.} Can the induced sense pathway serve as an alignment anchor for cross-lingual adaptation?

\paragraph{Methodology.} SENSIA \citep{cruz-etal-2026-multilinguality} tests whether the ACROS interface can serve as adaptation infrastructure. Starting from English ACROS, we adapt to Indonesian, Estonian, Swahili, and Turkish with the same English--target parallel pairs used in \citet{cruz-etal-2026-multilinguality}, comparing unadapted ACROS and Frozen BP repros. Intrinsic metrics are bidirectional contextual retrieval R@1, sense retrieval R@1 using SENSIA sense pooling, and target-language FLORES PPL.

We also run likelihood-based downstream tasks (XCOPA \citep{ponti-etal-2020-xcopa}, XStoryCloze \citep{lin-etal-2022-shot}, Belebele \citep{bandarkar-etal-2024-belebele}) and XL-Sum generation \citep{hasan-etal-2021-xl} on full test sets, with Gemma 3 270M \citep{gemma-team-2025-gemma3} and Qwen2 0.5B \citep{yang-etal-2024-qwen2} as dense multilingual baselines in the same parameter regime. Appendix~\ref{app:sensia} gives extra practical details.

\begin{table}[t]
  \centering
  {\small
  \setlength{\tabcolsep}{3pt}
  \begin{tabular}{lccc}
    \toprule
    Model & ctx@1 & sns@1 & PPL \\
    \midrule
    Unadapted & .009\tiny{ [.006,.011]} & .022\tiny{ [.018,.026]} & 107.7\tiny{ [105.6,110.0]} \\
    Frozen BP & .693\tiny{ [.680,.705]} & .558\tiny{ [.544,.571]} & 79.3\tiny{ [77.7,80.9]} \\
    ACROS & .988\tiny{ [.985,.991]} & .987\tiny{ [.984,.990]} & 7.94\tiny{ [7.84,8.05]} \\
    \bottomrule
  \end{tabular}}
  \caption{\textbf{For alignment, SENSIA uses the same induced sense path as a cross-lingual anchor.} After SENSIA adaptation from English to four languages, context- and sense-embedding based retrieval reach near-perfect R@1 on parallel FLORES pairs while target-language FLORES PPL drops sharply; intervals are language-balanced 95\% bootstraps over sentence pairs.}
  \label{tab:sensia}
\end{table}

\begin{table*}[t]
  \centering
  \small
  \setlength{\tabcolsep}{4pt}
  \begin{tabular*}{\textwidth}{@{}l@{\extracolsep{\fill}}rrrrrrrrr@{}}
    \toprule
    & \multicolumn{4}{c}{Likelihood acc.} & \multicolumn{5}{c}{XL-Sum gen.} \\
    \cmidrule(lr){2-5}\cmidrule(l){6-10}
    Model & XCOPA & XStory & Bele. & Mean & ROUGE-L & chrF++ & Empty & Rep-3g & Copy-4g \\
    \midrule
    Gemma 270M & 0.548 & 0.527 & 0.255 & 0.426 & \textbf{0.139} & \textbf{25.73} & 0.050 & \textbf{0.097} & 0.784 \\
    Qwen2 0.5B & 0.532 & 0.519 & \textbf{0.299} & \textbf{0.436} & 0.109 & 20.23 & 0.046 & 0.274 & 0.411 \\
    \midrule
    Base SmolLM2 & 0.520 & 0.492 & 0.239 & 0.402 & 0.088 & 15.30 & 0.008 & 0.359 & 0.368 \\
    Frozen BP & 0.530 & 0.485 & 0.242 & 0.405 & 0.013 & 2.81 & \textbf{0.000} & 0.454 & \textbf{0.000} \\
    ACROS & \textbf{0.558} & \textbf{0.542} & 0.231 & 0.424 & 0.110 & 18.30 & \textbf{0.000} & 0.283 & 0.125 \\
    \bottomrule
  \end{tabular*}
  \caption{\textbf{ACROS sense representations support effective adaptation of an English LM to other languages via SENSIA.} Macro-averaged downstream and generation results. Higher is better except Empty, Rep-3g, Copy-4g. Best scores in bold.}
  \label{tab:sensia-downstream}
\end{table*}

\begin{figure}[t]
  \centering
  \small
  \fbox{\begin{minipage}{0.95\columnwidth}
    \raggedright
    \textbf{Article:} Misa penahbisan Paus Fransiskus digelar \\[4pt]
    \textbf{Reference:} Misa penahbisan Paus Fransiskus diadakan di Lapangan Santo Petrus, Roma, di hadapan ratusan ribu orang, termasuk pemimpin dunia dan pemuka agama. \\[6pt]
    \textbf{Base SmolLM2:} \\
    \texttt{1.000.000 / Pesan: 1.000.000 / Pesan: 1.000.000 / Pesan: 1.000.000 / Pesan: ...} \\[4pt]
    {\footnotesize R-L: 0.000 \quad Copy-4g: 0.000 \quad Rep-3g: 0.778} \\[6pt]
    \textbf{ACROS + SENSIA:} \\
    Paus Fransiskus menyambut warga dan memuliakan Santo Petrus. Paus Fransiskus menyambut para anak-anak di Lapangan Santo Petrus. Paus Fransiskus menyambut para pemimpin negara dan tokoh politik dari berbagai negara. \\[4pt]
    {\footnotesize R-L: 0.308 \quad Copy-4g: 0.107 \quad Rep-3g: 0.172} \\[6pt]
    \textbf{Gemma 3 270M:} \\
    Paus Fransiskus menjalani serangkaian prosesi dalam penahbisannya. Paus meninggalkan kediaman sementaranya di Casa Santa Marta sebelum pukul 0900 waktu setempat. Sebelum menghadiri misa, Paus Fransiskus menyambut massa yang bersorak-sorai dan melambaikan bendera. \\[4pt]
    {\footnotesize R-L: 0.133 \quad Copy-4g: 1.000 \quad Rep-3g: 0.000}
  \end{minipage}}
  \caption{\textbf{ACROS-induced senses provide strong alignment anchors, allowing target-language summarization after SENSIA adaptation.} Base SmolLM2 fails to generate coherent summaries before adaptation. ACROS+SENSIA produces a coherent target-language summary with low source copying (Copy-4g 0.107) after SENSIA adaptation. This contrasts with Gemma, which is pretrained with a multilingual mix but copies almost entirely from the source (Copy-4g 1.000).}
  \label{fig:xlsum-qualitative}
\end{figure}

\paragraph{Results.} Table~\ref{tab:sensia} shows strong SENSIA alignment from ACROS senses. Before adaptation, ACROS has essentially no cross-lingual signal (ctx@1 = 0.009). After adaptation with English--target parallel pairs from \citet{cruz-etal-2026-multilinguality}, retrieval jumps to 0.988 contextual and 0.987 sense R@1, while target-language perplexity drops to 7.94. The Frozen BP control improves over the unadapted baseline but its degraded LM substrate limits both retrieval and target-language modeling.

Table~\ref{tab:sensia-downstream} shows downstream gains after SENSIA adaptation. ACROS improves over its base model on macro mean and on XCOPA and XStoryCloze, and is competitive with Gemma and Qwen despite starting as English-only. In generation, Gemma and Qwen score higher on lexical overlap but copy far more from the source (Copy-4g 0.784 and 0.411 versus 0.125 for ACROS). Appendices~\ref{app:downstream-cis}, \ref{app:prompt-continuations}, and~\ref{app:xlsum-qualitative} give per-language intervals, sample prompt continuations, and XL-Sum generations.

\section{Ablations}
\label{sec:ablations}

\subsection{Sense Interface vs. Dense Hidden States}

\paragraph{Research Question.} Does the explicit sense interface add value beyond the semantic information already present in the base model's dense hidden states?

\paragraph{Methodology.} We use two paired controls. For WSD, dense hidden states from the same SmolLM2 backbone replace ACROS sense activations in the same context--gloss matching protocol. For steering, a generous dense target-best control searches for a hidden coordinate/sign that maximizes CoInCo substitute mass under a norm budget matched to the selected sense intervention.

\begin{table}[t]
  \centering
  {\small
  \begin{tabular}{lrr}
    \toprule
    Probe & Dense/base & ACROS sense \\
    \midrule
    WSD F1 & 63.75 & \textbf{64.95} \\
    CoInCo delta & \textbf{+0.010240} & +0.001642 \\
    CoInCo KL & 1.77e-5 & \textbf{5.69e-7} \\
    \bottomrule
  \end{tabular}}
  \caption{\textbf{Dense hidden states already contain semantic signal, but ACROS makes that signal cleaner to access and reuse.} Dense target-best can push CoInCo substitute mass more strongly, but ACROS sense target-best achieves much lower KL and is semantically indexed, supporting the claim that ACROS changes access to meaning rather than inventing it from scratch.}
  \label{tab:hidden}
\end{table}

\begin{table}[t]
  \centering
  {\small
  \setlength{\tabcolsep}{3pt}
  \begin{tabularx}{\columnwidth}{@{}lccX@{}}
    \toprule
    Handle & Delta & KL & Largest content-word increases \\
    \midrule
    ACROS sense 19 & +0.003057 & 6.54e-7 &
    expense +0.0028 pp; burden +0.0013 pp; cost +0.0008 pp \\
    Dense coord. 87 & +0.021610 & 2.92e-5 &
    expense +0.0211 pp; burden +0.0118 pp; deficit +0.0058 pp \\
    \bottomrule
  \end{tabularx}}
  \caption{\textbf{Dense hidden states can steer strongly, but ACROS senses steer more surgically with much lower KL.} On \texttt{coinco:7535} (target word \texttt{charge} with human substitutes \texttt{expenditure, fee, cost, expense, amount, fine, obligation}), dense target-best moves substitute mass at about $45\times$ the KL with larger collateral shifts than ACROS. ``pp'' denotes probability points.}
  \label{tab:hidden-case}
\end{table}

\paragraph{Results.} Pretrained LMs already encode substantial semantic structure, and our ablation confirms this: dense hidden-state gloss matching reaches 63.75 F1, only 1.20 below ACROS but reliably different (McNemar $p=1.31\mathrm{e}{-9}$). What ACROS adds is \emph{not} the semantic information itself, but \emph{a cleaner way to access it}. On CoInCo, dense target-best pushes substitute mass about 6.2$\times$ harder, but at about 31$\times$ the KL. The single-row view in Table~\ref{tab:hidden-case} makes the tradeoff vivid: for \texttt{charge}, dense steering moves the right words but also moves many wrong ones, while ACROS sense 19 produces a smaller, more surgical shift. The explicit sense interface provides lower-dimensional, semantically indexed handles; crucially, these same handles also support WSD and cross-lingual alignment, which raw hidden-state coordinates do not. A residual addition without sense decomposition could add capacity, but would not produce individually addressable sense axes: the per-axis structure is what makes gloss matching, single-sense steering, and SENSIA sense-level pooling possible through the same set of variables.

Appendix~\ref{app:capacity-scale} reports additional ablations exploring sense-slot capacity and backbone scale.

\subsection{Architecture Portability}
\label{sec:portability}

\paragraph{Research Question.} Does ACROS induce a functional sense interface on decoder architectures beyond the Llama family?

\paragraph{Methodology.} We apply the same induction objective and hyperparameters to two additional backbones that differ from SmolLM2 in normalization, activation, position encoding, and block structure: Pythia-410M (parallel attention + FFN, LayerNorm, RoPE, GELU) and OPT-350M (sequential blocks, LayerNorm, learned positional embeddings, ReLU). All three models are induced on the same FineWeb English data and evaluated on FLORES PPL preservation, tokenizer-specific LAMA@1, WSD, and CoInCo steering under both oracle and non-oracle selectors.

\paragraph{Results.} Table~\ref{tab:portability} shows that ACROS induces functional sense interfaces on all three architectures. WSD is the cleanest result: Pythia essentially matches SmolLM2 (64.99 vs.\ 64.95 F1), and OPT is only half a point lower (64.46). Oracle steering reaches 100\% success ($\Delta>0$) on every backbone. The non-oracle self top-$k$ proxy achieves 81--90\% success rate ($\Delta>0$), while contribution-norm, random, and norm controls remain near chance on all three. LM behavior is preserved throughout, with small directional gains in both PPL and LAMA@1 within every backbone. Steering magnitudes are smaller on Pythia and OPT than on SmolLM2, so we frame this as portability of the interface rather than identical strength across architectures, but the core finding is that the same induction objective, applied without per-backbone tuning, produces a working sense interface on three architecturally distinct decoders. Additional ablations for larger backbone sizes and diagnostic runs are reported in Appendix~\ref{app:capacity-scale}.

\begin{table}[t]
  \centering
  {\small
  \begin{tabular}{lccc}
    \toprule
    Metric & SmolLM2 & Pythia & OPT \\
    \midrule
    FLORES PPL $\Delta$ & -0.05 & -0.35 & -0.62 \\
    LAMA@1 $\Delta$ & +.002 & +.010 & +.003 \\
    WSD F1 & 64.95 & 64.99 & 64.46 \\
    CoInCo oracle succ. & 100\% & 100\% & 100\% \\
    CoInCo self top-$k$ succ. & 89.8\% & 81.3\% & 85.4\% \\
    \bottomrule
  \end{tabular}}
  \caption{\textbf{ACROS induces functional sense interfaces across architecturally distinct decoders.} SmolLM2 (Llama-style), Pythia (parallel blocks, GELU), and OPT (learned positions, ReLU) all receive the same induction objective and hyperparameters. $\Delta$ rows show ACROS minus the original backbone (lower PPL and higher LAMA@1 are better). PPL and LAMA use each backbone's own tokenizer and are comparable within rows but not across them. Success is the fraction of ${\sim}5{,}160$ scored cases where $\Delta>0$.}
  \label{tab:portability}
\end{table}

\section{Related Work}
\label{sec:related}

Several recent methods add semantic structure to decoder models. Multi-Sense Embeddings \citep{wang-etal-2025-multi-sense}, DSKD \citep{wang-etal-2026-dskd}, CoCoMix \citep{tack-etal-2026-cocomix}, and codebook features \citep{pmlr-v235-tamkin24a} all build or inject explicit semantic variables during training, but none are designed as a single inference-time sense interface that supports measurement, steering, and alignment through the same variables. Probing work has long shown that pretrained LMs encode word-sense information internally; our Section~\ref{sec:ablations} ablation confirms this and shows that ACROS surfaces it as a cleaner, lower-KL interface rather than creating it from scratch. Post-hoc steering via activation additions \citep{turner-etal-2023-activation}, representation engineering \citep{zou-etal-2023-representation}, or SAE features \citep{rimsky-etal-2024-steering,arad-etal-2025-saes,chalnev-etal-2024-improving} can modify behavior but is often brittle \citep{pmlr-v267-wu25a,silva-etal-2025-steering}. ACROS provides trained-in, token-local sense slots that give reliable low-KL handles for word-level steering.

\section{Conclusion}
\label{sec:conclusion}

We have shown that sense representations do not need to be baked into a model's architecture. ACROS induces an explicit, gated sense pathway into an ordinary pretrained decoder LM, and the same set of induced variables supports three capabilities that previously required separate systems: zero-shot word-sense disambiguation competitive with the WordNet first-sense heuristic, low-KL lexical steering where a simple non-oracle semantic proxy recovers about 90\% of positive substitute-mass shifts across 5,161 cases, and SENSIA cross-lingual alignment that reaches near-perfect retrieval while reducing target-language perplexity by an order of magnitude over Backpack conversion. The residual design preserves the base model's prediction quality throughout; ACROS adds structure without taking anything away. These results establish that sense-level measurement, intervention, and alignment are \emph{inducible} properties: they can be added to an existing LM rather than requiring a purpose-built architecture.

\section*{Limitations}

\paragraph{Lexical steering scope.} Steering is evaluated via a deliberate local intervention: a $1.2\times$ boost on a single sense at a single position, prioritizing low KL over large probability shifts. An oracle selector using gold substitute access establishes that useful causal directions exist; a non-oracle semantic proxy (self top-$k$) recovers 81--90\% of oracle success without gold access, while magnitude-only controls remain near chance. A learned selector that replaces the proxy is left for future work.

\paragraph{WSD scope.} WSD is evaluated zero-shot using the same induced variables that also support steering and alignment; no sense supervision is used during induction. ACROS reaches 64.95 F1 on Raganato ALL, competitive with the WordNet first-sense heuristic (65.2). Supervised systems score higher, as expected, but operate on dedicated architectures that do not simultaneously support steering and cross-lingual alignment.

\paragraph{Cross-lingual evaluation scope.} Cross-lingual evaluation uses SENSIA \citep{cruz-etal-2026-multilinguality}, a method architecturally compatible with ACROS, which benefits intrinsic retrieval. Downstream results show that sense-based alignment transfers effectively to lexical and short-context tasks (XCOPA, XStoryCloze); passage-level reasoning (Belebele) transfers less well, and natively multilingual pretraining remains better suited for that setting. Broader evaluation with other adaptation methods and languages is left for future work.

\paragraph{Architecture and scale.} Three decoder families are tested (Llama, GPT-NeoX, OPT); encoder-decoders and models above 2B remain future work. Additional ablations for larger backbone sizes are available in Appendix~\ref{app:capacity-scale}, characterizing how the current induction recipe behaves at scale.

\paragraph{English-only induction.} Cross-lingual ability is scoped to post-hoc SENSIA adaptation from an English-only induced interface; multilingual induction is out of scope and left for future work.

\paragraph{Automatic evaluation only.} All evaluations use automatic metrics; no formal human judgments are reported.

\section*{Ethical Considerations}

This work uses public benchmark and training resources and does not introduce new human-subject data collection. The main ethical risk is inherited from the pretrained base models and datasets: ACROS can expose and steer lexical associations that may encode social biases, stereotypes, or uneven coverage across languages and communities.

The steering results should be read as diagnostic evidence about local lexical control, not as a deployment-ready editing system. Because low-KL interventions can subtly change next-token preferences, similar mechanisms could be misused for lexical framing or manipulation if paired with an improved selector. The self top-$k$ proxy demonstrates that partial non-oracle selection is feasible, which makes the misuse risk more concrete than a purely oracle result; we note this as a reason for caution in downstream applications.

The cross-lingual adaptation experiments cover Indonesian, Estonian, Swahili, and Turkish, but the evaluations are automatic and limited. We do not claim that the adapted models are robust, culturally appropriate, or production-ready for speakers of those languages. Any release or downstream use should include language-community evaluation, bias testing, and clear disclosure that ACROS is a research interface-induction method.

\paragraph{AI use disclaimer.} Anthropic's Claude was used to assist with proofreading and editing the final draft of this paper.

\bibliography{custom}

@inproceedings{
    allal-etal-2025-smollm2,
    title={Smol{LM}2: When Smol Goes Big {\textemdash} Data-Centric Training of a Fully Open Small Language Model},
    author={Loubna Ben Allal and Anton Lozhkov and Elie Bakouch and Gabriel Martin Blazquez and Guilherme Penedo and Lewis Tunstall and Andr{\'e}s Marafioti and Agust{\'\i}n Piqueres Lajar{\'\i}n and Hynek Kydl{\'\i}{\v{c}}ek and Vaibhav Srivastav and Joshua Lochner and Caleb Fahlgren and Xuan Son Nguyen and Ben Burtenshaw and Cl{\'e}mentine Fourrier and Haojun Zhao and Hugo Larcher and Mathieu Morlon and Cyril Zakka and Colin Raffel and Leandro Von Werra and Thomas Wolf},
    booktitle={Second Conference on Language Modeling},
    year={2025},
    url={https://openreview.net/forum?id=3JiCl2A14H}
}

@inproceedings{arad-etal-2025-saes,
    title = "{SAE}s Are Good for Steering {--} If You Select the Right Features",
    author = "Arad, Dana  and
      Mueller, Aaron  and
      Belinkov, Yonatan",
    editor = "Christodoulopoulos, Christos  and
      Chakraborty, Tanmoy  and
      Rose, Carolyn  and
      Peng, Violet",
    booktitle = "Proceedings of the 2025 Conference on Empirical Methods in Natural Language Processing",
    month = nov,
    year = "2025",
    address = "Suzhou, China",
    publisher = "Association for Computational Linguistics",
    url = "https://aclanthology.org/2025.emnlp-main.519/",
    doi = "10.18653/v1/2025.emnlp-main.519",
    pages = "10241--10259",
    ISBN = "979-8-89176-332-6"
}

@inproceedings{bandarkar-etal-2024-belebele,
    title = "The Belebele Benchmark: a Parallel Reading Comprehension Dataset in 122 Language Variants",
    author = "Bandarkar, Lucas  and
      Liang, Davis  and
      Muller, Benjamin  and
      Artetxe, Mikel  and
      Shukla, Satya Narayan  and
      Husa, Donald  and
      Goyal, Naman  and
      Krishnan, Abhinandan  and
      Zettlemoyer, Luke  and
      Khabsa, Madian",
    editor = "Ku, Lun-Wei  and
      Martins, Andre  and
      Srikumar, Vivek",
    booktitle = "Proceedings of the 62nd Annual Meeting of the Association for Computational Linguistics (Volume 1: Long Papers)",
    month = aug,
    year = "2024",
    address = "Bangkok, Thailand",
    publisher = "Association for Computational Linguistics",
    url = "https://aclanthology.org/2024.acl-long.44/",
    doi = "10.18653/v1/2024.acl-long.44",
    pages = "749--775"
}

@inproceedings{blevins-zettlemoyer-2020-moving,
    title = "Moving Down the Long Tail of Word Sense Disambiguation with Gloss Informed Bi-encoders",
    author = "Blevins, Terra  and
      Zettlemoyer, Luke",
    editor = "Jurafsky, Dan  and
      Chai, Joyce  and
      Schluter, Natalie  and
      Tetreault, Joel",
    booktitle = "Proceedings of the 58th Annual Meeting of the Association for Computational Linguistics",
    month = jul,
    year = "2020",
    address = "Online",
    publisher = "Association for Computational Linguistics",
    url = "https://aclanthology.org/2020.acl-main.95/",
    doi = "10.18653/v1/2020.acl-main.95",
    pages = "1006--1017"
}

@misc{chalnev-etal-2024-improving,
      title={Improving Steering Vectors by Targeting Sparse Autoencoder Features}, 
      author={Sviatoslav Chalnev and Matthew Siu and Arthur Conmy},
      year={2024},
      eprint={2411.02193},
      archivePrefix={arXiv},
      primaryClass={cs.LG},
      url={https://arxiv.org/abs/2411.02193}, 
}

@misc{cruz-etal-2026-multilinguality,
      title={Multilinguality as Sense Adaptation}, 
      author={Jan Christian Blaise Cruz and David Ifeoluwa Adelani and Alham Fikri Aji},
      year={2026},
      eprint={2601.10310},
      archivePrefix={arXiv},
      primaryClass={cs.CL},
      url={https://arxiv.org/abs/2601.10310}, 
}

@inproceedings{silva-etal-2025-steering,
    title = "Steering off Course: Reliability Challenges in Steering Language Models",
    author = "Da Silva, Patrick Queiroz  and
      Sethuraman, Hari  and
      Rajagopal, Dheeraj  and
      Hajishirzi, Hannaneh  and
      Kumar, Sachin",
    editor = "Che, Wanxiang  and
      Nabende, Joyce  and
      Shutova, Ekaterina  and
      Pilehvar, Mohammad Taher",
    booktitle = "Proceedings of the 63rd Annual Meeting of the Association for Computational Linguistics (Volume 1: Long Papers)",
    month = jul,
    year = "2025",
    address = "Vienna, Austria",
    publisher = "Association for Computational Linguistics",
    url = "https://aclanthology.org/2025.acl-long.974/",
    doi = "10.18653/v1/2025.acl-long.974",
    pages = "19856--19882",
    ISBN = "979-8-89176-251-0"
}

@inproceedings{dathathri-etal-2020-plug,
title={Plug and Play Language Models: A Simple Approach to Controlled Text Generation},
author={Sumanth Dathathri and Andrea Madotto and Janice Lan and Jane Hung and Eric Frank and Piero Molino and Jason Yosinski and Rosanne Liu},
booktitle={International Conference on Learning Representations},
year={2020},
url={https://openreview.net/forum?id=H1edEyBKDS}
}

@inproceedings{finkelstein-etal-2001-placing,
  author    = {Finkelstein, Lev and Gabrilovich, Evgeniy and Matias, Yossi and
               Rivlin, Ehud and Solan, Zach and Wolfman, Gadi and Ruppin, Eytan},
  title     = {Placing Search in Context: The Concept Revisited},
  booktitle = {Proceedings of the 10th International Conference on the World Wide Web},
  pages     = {406--414},
  year      = {2001},
  publisher = {ACM}
}

@misc{gemma-team-2025-gemma3,
      title={Gemma 3 Technical Report}, 
      author={{Gemma Team} and Aishwarya Kamath and Johan Ferret and Shreya Pathak and Nino Vieillard and Ramona Merhej and Sarah Perrin and Tatiana Matejovicova and Alexandre Ramé and Morgane Rivière and Louis Rouillard and Thomas Mesnard and Geoffrey Cideron and Jean-bastien Grill and Sabela Ramos and Edouard Yvinec and Michelle Casbon and Etienne Pot and Ivo Penchev and Gaël Liu and Francesco Visin and Kathleen Kenealy and Lucas Beyer and Xiaohai Zhai and Anton Tsitsulin and Robert Busa-Fekete and Alex Feng and Noveen Sachdeva and Benjamin Coleman and Yi Gao and Basil Mustafa and Iain Barr and Emilio Parisotto and David Tian and Matan Eyal and Colin Cherry and Jan-Thorsten Peter and Danila Sinopalnikov and Surya Bhupatiraju and Rishabh Agarwal and Mehran Kazemi and Dan Malkin and Ravin Kumar and David Vilar and Idan Brusilovsky and Jiaming Luo and Andreas Steiner and Abe Friesen and Abhanshu Sharma and Abheesht Sharma and Adi Mayrav Gilady and Adrian Goedeckemeyer and Alaa Saade and Alex Feng and Alexander Kolesnikov and Alexei Bendebury and Alvin Abdagic and Amit Vadi and András György and André Susano Pinto and Anil Das and Ankur Bapna and Antoine Miech and Antoine Yang and Antonia Paterson and Ashish Shenoy and Ayan Chakrabarti and Bilal Piot and Bo Wu and Bobak Shahriari and Bryce Petrini and Charlie Chen and Charline Le Lan and Christopher A. Choquette-Choo and CJ Carey and Cormac Brick and Daniel Deutsch and Danielle Eisenbud and Dee Cattle and Derek Cheng and Dimitris Paparas and Divyashree Shivakumar Sreepathihalli and Doug Reid and Dustin Tran and Dustin Zelle and Eric Noland and Erwin Huizenga and Eugene Kharitonov and Frederick Liu and Gagik Amirkhanyan and Glenn Cameron and Hadi Hashemi and Hanna Klimczak-Plucińska and Harman Singh and Harsh Mehta and Harshal Tushar Lehri and Hussein Hazimeh and Ian Ballantyne and Idan Szpektor and Ivan Nardini and Jean Pouget-Abadie and Jetha Chan and Joe Stanton and John Wieting and Jonathan Lai and Jordi Orbay and Joseph Fernandez and Josh Newlan and Ju-yeong Ji and Jyotinder Singh and Kat Black and Kathy Yu and Kevin Hui and Kiran Vodrahalli and Klaus Greff and Linhai Qiu and Marcella Valentine and Marina Coelho and Marvin Ritter and Matt Hoffman and Matthew Watson and Mayank Chaturvedi and Michael Moynihan and Min Ma and Nabila Babar and Natasha Noy and Nathan Byrd and Nick Roy and Nikola Momchev and Nilay Chauhan and Noveen Sachdeva and Oskar Bunyan and Pankil Botarda and Paul Caron and Paul Kishan Rubenstein and Phil Culliton and Philipp Schmid and Pier Giuseppe Sessa and Pingmei Xu and Piotr Stanczyk and Pouya Tafti and Rakesh Shivanna and Renjie Wu and Renke Pan and Reza Rokni and Rob Willoughby and Rohith Vallu and Ryan Mullins and Sammy Jerome and Sara Smoot and Sertan Girgin and Shariq Iqbal and Shashir Reddy and Shruti Sheth and Siim Põder and Sijal Bhatnagar and Sindhu Raghuram Panyam and Sivan Eiger and Susan Zhang and Tianqi Liu and Trevor Yacovone and Tyler Liechty and Uday Kalra and Utku Evci and Vedant Misra and Vincent Roseberry and Vlad Feinberg and Vlad Kolesnikov and Woohyun Han and Woosuk Kwon and Xi Chen and Yinlam Chow and Yuvein Zhu and Zichuan Wei and Zoltan Egyed and Victor Cotruta and Minh Giang and Phoebe Kirk and Anand Rao and Kat Black and Nabila Babar and Jessica Lo and Erica Moreira and Luiz Gustavo Martins and Omar Sanseviero and Lucas Gonzalez and Zach Gleicher and Tris Warkentin and Vahab Mirrokni and Evan Senter and Eli Collins and Joelle Barral and Zoubin Ghahramani and Raia Hadsell and Yossi Matias and D. Sculley and Slav Petrov and Noah Fiedel and Noam Shazeer and Oriol Vinyals and Jeff Dean and Demis Hassabis and Koray Kavukcuoglu and Clement Farabet and Elena Buchatskaya and Jean-Baptiste Alayrac and Rohan Anil and Dmitry and Lepikhin and Sebastian Borgeaud and Olivier Bachem and Armand Joulin and Alek Andreev and Cassidy Hardin and Robert Dadashi and Léonard Hussenot},
      year={2025},
      eprint={2503.19786},
      archivePrefix={arXiv},
      primaryClass={cs.CL},
      url={https://arxiv.org/abs/2503.19786}, 
}

@article{goyal-etal-2022-flores,
    author = {Goyal, Naman and Gao, Cynthia and Chaudhary, Vishrav and Chen, Peng-Jen and Wenzek, Guillaume and Ju, Da and Krishnan, Sanjana and Ranzato, Marc’Aurelio and Guzmán, Francisco and Fan, Angela},
    title = {The Flores-101 Evaluation Benchmark for Low-Resource and Multilingual Machine Translation},
    journal = {Transactions of the Association for Computational Linguistics},
    volume = {10},
    pages = {522-538},
    year = {2022},
    month = {05},
    issn = {2307-387X},
    doi = {10.1162/tacl_a_00474},
    url = {https://doi.org/10.1162/tacl_a_00474},
    eprint = {https://direct.mit.edu/tacl/article-pdf/doi/10.1162/tacl_a_00474/2020699/tacl_a_00474.pdf},
}

@inproceedings{hasan-etal-2021-xl,
    title = "{XL}-Sum: Large-Scale Multilingual Abstractive Summarization for 44 Languages",
    author = "Hasan, Tahmid  and
      Bhattacharjee, Abhik  and
      Islam, Md. Saiful  and
      Mubasshir, Kazi  and
      Li, Yuan-Fang  and
      Kang, Yong-Bin  and
      Rahman, M. Sohel  and
      Shahriyar, Rifat",
    editor = "Zong, Chengqing  and
      Xia, Fei  and
      Li, Wenjie  and
      Navigli, Roberto",
    booktitle = "Findings of the Association for Computational Linguistics: ACL-IJCNLP 2021",
    month = aug,
    year = "2021",
    address = "Online",
    publisher = "Association for Computational Linguistics",
    url = "https://aclanthology.org/2021.findings-acl.413/",
    doi = "10.18653/v1/2021.findings-acl.413",
    pages = "4693--4703"
}

@misc{hewitt-etal-2024-model-editing,
      title={Model Editing with Canonical Examples}, 
      author={John Hewitt and Sarah Chen and Lanruo Lora Xie and Edward Adams and Percy Liang and Christopher D. Manning},
      year={2024},
      eprint={2402.06155},
      archivePrefix={arXiv},
      primaryClass={cs.CL},
      url={https://arxiv.org/abs/2402.06155}, 
}

@inproceedings{hewitt-etal-2023-backpack,
    title = "Backpack Language Models",
    author = "Hewitt, John  and
      Thickstun, John  and
      Manning, Christopher  and
      Liang, Percy",
    editor = "Rogers, Anna  and
      Boyd-Graber, Jordan  and
      Okazaki, Naoaki",
    booktitle = "Proceedings of the 61st Annual Meeting of the Association for Computational Linguistics (Volume 1: Long Papers)",
    month = jul,
    year = "2023",
    address = "Toronto, Canada",
    publisher = "Association for Computational Linguistics",
    url = "https://aclanthology.org/2023.acl-long.506/",
    doi = "10.18653/v1/2023.acl-long.506",
    pages = "9103--9125"
}

@misc{keskar-etal-2019-ctrl,
      title={CTRL: A Conditional Transformer Language Model for Controllable Generation}, 
      author={Nitish Shirish Keskar and Bryan McCann and Lav R. Varshney and Caiming Xiong and Richard Socher},
      year={2019},
      eprint={1909.05858},
      archivePrefix={arXiv},
      primaryClass={cs.CL},
      url={https://arxiv.org/abs/1909.05858}, 
}

@inproceedings{kremer-etal-2014-substitutes,
    title = "What Substitutes Tell Us - Analysis of an ``All-Words'' Lexical Substitution Corpus",
    author = "Kremer, Gerhard  and
      Erk, Katrin  and
      Pad{\'o}, Sebastian  and
      Thater, Stefan",
    editor = "Wintner, Shuly  and
      Goldwater, Sharon  and
      Riezler, Stefan",
    booktitle = "Proceedings of the 14th Conference of the {E}uropean Chapter of the Association for Computational Linguistics",
    month = apr,
    year = "2014",
    address = "Gothenburg, Sweden",
    publisher = "Association for Computational Linguistics",
    url = "https://aclanthology.org/E14-1057/",
    doi = "10.3115/v1/E14-1057",
    pages = "540--549"
}

@inproceedings{lin-etal-2022-shot,
    title = "Few-shot Learning with Multilingual Generative Language Models",
    author = "Lin, Xi Victoria  and
      Mihaylov, Todor  and
      Artetxe, Mikel  and
      Wang, Tianlu  and
      Chen, Shuohui  and
      Simig, Daniel  and
      Ott, Myle  and
      Goyal, Naman  and
      Bhosale, Shruti  and
      Du, Jingfei  and
      Pasunuru, Ramakanth  and
      Shleifer, Sam  and
      Koura, Punit Singh  and
      Chaudhary, Vishrav  and
      O{'}Horo, Brian  and
      Wang, Jeff  and
      Zettlemoyer, Luke  and
      Kozareva, Zornitsa  and
      Diab, Mona  and
      Stoyanov, Veselin  and
      Li, Xian",
    editor = "Goldberg, Yoav  and
      Kozareva, Zornitsa  and
      Zhang, Yue",
    booktitle = "Proceedings of the 2022 Conference on Empirical Methods in Natural Language Processing",
    month = dec,
    year = "2022",
    address = "Abu Dhabi, United Arab Emirates",
    publisher = "Association for Computational Linguistics",
    url = "https://aclanthology.org/2022.emnlp-main.616/",
    doi = "10.18653/v1/2022.emnlp-main.616",
    pages = "9019--9052"
}

@article{loureiro-etal-2022-lmms,
    title = {LMMS reloaded: Transformer-based sense embeddings for disambiguation and beyond},
    journal = {Artificial Intelligence},
    volume = {305},
    pages = {103661},
    year = {2022},
    issn = {0004-3702},
    doi = {https://doi.org/10.1016/j.artint.2022.103661},
    url = {https://www.sciencedirect.com/science/article/pii/S0004370222000017},
    author = {Daniel Loureiro and Alípio {Mário Jorge} and Jose Camacho-Collados}
}

@inproceedings{petroni-etal-2019-language,
    title = "Language Models as Knowledge Bases?",
    author = {Petroni, Fabio  and
      Rockt{\"a}schel, Tim  and
      Riedel, Sebastian  and
      Lewis, Patrick  and
      Bakhtin, Anton  and
      Wu, Yuxiang  and
      Miller, Alexander},
    editor = "Inui, Kentaro  and
      Jiang, Jing  and
      Ng, Vincent  and
      Wan, Xiaojun",
    booktitle = "Proceedings of the 2019 Conference on Empirical Methods in Natural Language Processing and the 9th International Joint Conference on Natural Language Processing (EMNLP-IJCNLP)",
    month = nov,
    year = "2019",
    address = "Hong Kong, China",
    publisher = "Association for Computational Linguistics",
    url = "https://aclanthology.org/D19-1250/",
    doi = "10.18653/v1/D19-1250",
    pages = "2463--2473"
}

@inproceedings{ponti-etal-2020-xcopa,
    title = "{XCOPA}: A Multilingual Dataset for Causal Commonsense Reasoning",
    author = "Ponti, Edoardo Maria  and
      Glava{\v{s}}, Goran  and
      Majewska, Olga  and
      Liu, Qianchu  and
      Vuli{\'c}, Ivan  and
      Korhonen, Anna",
    editor = "Webber, Bonnie  and
      Cohn, Trevor  and
      He, Yulan  and
      Liu, Yang",
    booktitle = "Proceedings of the 2020 Conference on Empirical Methods in Natural Language Processing (EMNLP)",
    month = nov,
    year = "2020",
    address = "Online",
    publisher = "Association for Computational Linguistics",
    url = "https://aclanthology.org/2020.emnlp-main.185/",
    doi = "10.18653/v1/2020.emnlp-main.185",
    pages = "2362--2376"
}

@inproceedings{raganato-etal-2017-word,
    title = "Word Sense Disambiguation: A Unified Evaluation Framework and Empirical Comparison",
    author = "Raganato, Alessandro  and
      Camacho-Collados, Jose  and
      Navigli, Roberto",
    editor = "Lapata, Mirella  and
      Blunsom, Phil  and
      Koller, Alexander",
    booktitle = "Proceedings of the 15th Conference of the {E}uropean Chapter of the Association for Computational Linguistics: Volume 1, Long Papers",
    month = apr,
    year = "2017",
    address = "Valencia, Spain",
    publisher = "Association for Computational Linguistics",
    url = "https://aclanthology.org/E17-1010/",
    pages = "99--110"
}

@inproceedings{rimsky-etal-2024-steering,
    title = "Steering Llama 2 via Contrastive Activation Addition",
    author = "Rimsky, Nina  and
      Gabrieli, Nick  and
      Schulz, Julian  and
      Tong, Meg  and
      Hubinger, Evan  and
      Turner, Alexander",
    editor = "Ku, Lun-Wei  and
      Martins, Andre  and
      Srikumar, Vivek",
    booktitle = "Proceedings of the 62nd Annual Meeting of the Association for Computational Linguistics (Volume 1: Long Papers)",
    month = aug,
    year = "2024",
    address = "Bangkok, Thailand",
    publisher = "Association for Computational Linguistics",
    url = "https://aclanthology.org/2024.acl-long.828/",
    doi = "10.18653/v1/2024.acl-long.828",
    pages = "15504--15522"
}

@inproceedings{tack-etal-2026-cocomix,
    title={{LLM} Pretraining with Continuous Concepts},
    author={Jihoon Tack and Jack Lanchantin and Jane Yu and Andrew Cohen and Ilia Kulikov and Janice Lan and Shibo Hao and Yuandong Tian and Jason E Weston and Xian Li},
    booktitle={The Fourteenth International Conference on Learning Representations},
    year={2026},
    url={https://openreview.net/forum?id=wTGcb3DxOn}
}

@InProceedings{pmlr-v235-tamkin24a,
  title = 	 {Codebook Features: Sparse and Discrete Interpretability for Neural Networks},
  author =       {Tamkin, Alex and Taufeeque, Mohammad and Goodman, Noah},
  booktitle = 	 {Proceedings of the 41st International Conference on Machine Learning},
  pages = 	 {47535--47563},
  year = 	 {2024},
  editor = 	 {Salakhutdinov, Ruslan and Kolter, Zico and Heller, Katherine and Weller, Adrian and Oliver, Nuria and Scarlett, Jonathan and Berkenkamp, Felix},
  volume = 	 {235},
  series = 	 {Proceedings of Machine Learning Research},
  month = 	 {21--27 Jul},
  publisher =    {PMLR},
  url = 	 {https://proceedings.mlr.press/v235/tamkin24a.html}
}

@misc{turner-etal-2023-activation,
      title={Steering Language Models With Activation Engineering}, 
      author={Alexander Matt Turner and Lisa Thiergart and Gavin Leech and David Udell and Juan J. Vazquez and Ulisse Mini and Monte MacDiarmid},
      year={2024},
      eprint={2308.10248},
      archivePrefix={arXiv},
      primaryClass={cs.CL},
      url={https://arxiv.org/abs/2308.10248}, 
}

@inproceedings{wang-etal-2025-multi-sense,
    title = "Multi-Sense Embeddings for Language Models and Knowledge Distillation",
    author = "Wang, Qitong  and
      Zaki, Mohammed J  and
      Kollias, Georgios  and
      Kalantzis, Vasileios",
    editor = "Che, Wanxiang  and
      Nabende, Joyce  and
      Shutova, Ekaterina  and
      Pilehvar, Mohammad Taher",
    booktitle = "Findings of the Association for Computational Linguistics: ACL 2025",
    month = jul,
    year = "2025",
    address = "Vienna, Austria",
    publisher = "Association for Computational Linguistics",
    url = "https://aclanthology.org/2025.findings-acl.691/",
    doi = "10.18653/v1/2025.findings-acl.691",
    pages = "13353--13369",
    ISBN = "979-8-89176-256-5"
}

@misc{wang-etal-2026-dskd,
      title={Decoder-based Sense Knowledge Distillation}, 
      author={Qitong Wang and Mohammed J. Zaki and Georgios Kollias and Vasileios Kalantzis},
      year={2026},
      eprint={2602.22351},
      archivePrefix={arXiv},
      primaryClass={cs.CL},
      url={https://arxiv.org/abs/2602.22351}, 
}

@InProceedings{pmlr-v267-wu25a,
  title = 	 {{A}x{B}ench: Steering {LLM}s? {E}ven Simple Baselines Outperform Sparse Autoencoders},
  author =       {Wu, Zhengxuan and Arora, Aryaman and Geiger, Atticus and Wang, Zheng and Huang, Jing and Jurafsky, Dan and Manning, Christopher D and Potts, Christopher},
  booktitle = 	 {Proceedings of the 42nd International Conference on Machine Learning},
  pages = 	 {67035--67080},
  year = 	 {2025},
  editor = 	 {Singh, Aarti and Fazel, Maryam and Hsu, Daniel and Lacoste-Julien, Simon and Berkenkamp, Felix and Maharaj, Tegan and Wagstaff, Kiri and Zhu, Jerry},
  volume = 	 {267},
  series = 	 {Proceedings of Machine Learning Research},
  month = 	 {13--19 Jul},
  publisher =    {PMLR},
  url = 	 {https://proceedings.mlr.press/v267/wu25a.html}
}

@misc{yang-etal-2024-qwen2,
      title={Qwen2 Technical Report}, 
      author={An Yang and Baosong Yang and Binyuan Hui and Bo Zheng and Bowen Yu and Chang Zhou and Chengpeng Li and Chengyuan Li and Dayiheng Liu and Fei Huang and Guanting Dong and Haoran Wei and Huan Lin and Jialong Tang and Jialin Wang and Jian Yang and Jianhong Tu and Jianwei Zhang and Jianxin Ma and Jianxin Yang and Jin Xu and Jingren Zhou and Jinze Bai and Jinzheng He and Junyang Lin and Kai Dang and Keming Lu and Keqin Chen and Kexin Yang and Mei Li and Mingfeng Xue and Na Ni and Pei Zhang and Peng Wang and Ru Peng and Rui Men and Ruize Gao and Runji Lin and Shijie Wang and Shuai Bai and Sinan Tan and Tianhang Zhu and Tianhao Li and Tianyu Liu and Wenbin Ge and Xiaodong Deng and Xiaohuan Zhou and Xingzhang Ren and Xinyu Zhang and Xipin Wei and Xuancheng Ren and Xuejing Liu and Yang Fan and Yang Yao and Yichang Zhang and Yu Wan and Yunfei Chu and Yuqiong Liu and Zeyu Cui and Zhenru Zhang and Zhifang Guo and Zhihao Fan},
      year={2024},
      eprint={2407.10671},
      archivePrefix={arXiv},
      primaryClass={cs.CL},
      url={https://arxiv.org/abs/2407.10671}, 
}

@inproceedings{yang-klein-2021-fudge,
    title = "{FUDGE}: Controlled Text Generation With Future Discriminators",
    author = "Yang, Kevin  and
      Klein, Dan",
    editor = "Toutanova, Kristina  and
      Rumshisky, Anna  and
      Zettlemoyer, Luke  and
      Hakkani-Tur, Dilek  and
      Beltagy, Iz  and
      Bethard, Steven  and
      Cotterell, Ryan  and
      Chakraborty, Tanmoy  and
      Zhou, Yichao",
    booktitle = "Proceedings of the 2021 Conference of the North American Chapter of the Association for Computational Linguistics: Human Language Technologies",
    month = jun,
    year = "2021",
    address = "Online",
    publisher = "Association for Computational Linguistics",
    url = "https://aclanthology.org/2021.naacl-main.276/",
    doi = "10.18653/v1/2021.naacl-main.276",
    pages = "3511--3535"
}

@misc{zou-etal-2023-representation,
      title={Representation Engineering: A Top-Down Approach to AI Transparency}, 
      author={Andy Zou and Long Phan and Sarah Chen and James Campbell and Phillip Guo and Richard Ren and Alexander Pan and Xuwang Yin and Mantas Mazeika and Ann-Kathrin Dombrowski and Shashwat Goel and Nathaniel Li and Michael J. Byun and Zifan Wang and Alex Mallen and Steven Basart and Sanmi Koyejo and Dawn Song and Matt Fredrikson and J. Zico Kolter and Dan Hendrycks},
      year={2025},
      eprint={2310.01405},
      archivePrefix={arXiv},
      primaryClass={cs.LG},
      url={https://arxiv.org/abs/2310.01405}, 
}

\appendix

\section{Backpack Conversion Diagnostics}
\label{app:conversion}

We evaluate whether an existing pretrained decoder can be converted into a Backpack-style LM without losing ordinary language-model quality. The source model is \texttt{HuggingFaceTB/SmolLM2-360M}, an English-only Llama-style decoder. We compare the original model against three Backpack conversion families:

\begin{itemize}
  \item \textbf{Continued LM training:} initialize a Backpack-Llama model from the pretrained checkpoint and continue causal-LM training in the conversion-output architecture.
  \item \textbf{Distillation:} train the converted Backpack student with the pretrained SmolLM2 model as a frozen teacher, combining next-token training with teacher-student alignment.
  \item \textbf{Frozen-backbone distillation:} freeze the SmolLM2 backbone and LM head inside the Backpack student, and train only the sense-side modules. This tests whether the failure is ordinary backbone drift or the conversion-output family itself.
\end{itemize}

We use sentence-level FLORES English devtest perplexity and filtered LAMA factual cloze accuracy as compact LM-health diagnostics. Fixed generation smoke tests were also used to check whether fluent prediction behavior survived conversion.

\begin{table*}[t]
  \centering
  \small
  \begin{tabular}{lrrrr}
    \toprule
    Variant & FLORES PPL & LAMA@1 & LAMA@5 & MRR \\
    \midrule
    Original & 25.1 & 0.3155 & 0.525 & 0.4161 \\
    Backpack + CPT & 356.86 & 0.0695 & 0.146 & 0.1224 \\
    Backpack + Distil. & 196.30 & 0.0420 & 0.135 & 0.0926 \\
    Backpack + Frozen Distil. K8 & 825.98 & 0.0520 & 0.140 & 0.1034 \\
    Backpack + Distil. + Freeze & 236.94 & 0.1245 & 0.266 & 0.2004 \\
    \textbf{ACROS} & \textbf{25.09} & \textbf{0.3175} & \textbf{0.5245} & \textbf{0.4171} \\
    \bottomrule
  \end{tabular}
  \caption{Conversion diagnostics. Backpack conversion substantially degrades ordinary LM behavior, including under frozen-backbone controls. The residual ACROS path preserves base-level LM health.}
  \label{tab:conversion-diagnostics}
\end{table*}

The frozen-backbone controls are the key separation. If conversion failed only because the pretrained weights drifted during training, freezing the backbone and LM head should preserve base-model behavior. However, as the results show, it does not. Even with the original backbone and head frozen, the converted Backpack output remains much worse than the base LM. This supports the interpretation that the output family itself is the problem: the sense mixture is being asked to replace too much of the pretrained hidden state.

\section{SVD Bottleneck Diagnostic}
\label{app:svd}

The converted Backpack architecture routes prediction through a normalized sense mixture:
\begin{equation}
  h_q^{\mathrm{BP}}=\sum_{k=1}^{K}\alpha_{q,k}v_{q,k},
  \quad \alpha_{q,k}\ge0,\quad \sum_k\alpha_{q,k}=1.
\end{equation}
For each position, this representation lies in the convex hull of the active sense vectors before the LM head. This is useful when the model is pretrained in that form, but it is restrictive when the mixture is asked to replace the full hidden state of an existing LM.

We estimate the size of the base hidden-state space by collecting 10,000 final-layer hidden states from the pretrained SmolLM2-360M model on English FineWeb tokens. We center the hidden-state matrix, compute its singular values, and measure cumulative variance explained by the top ranks. The result is not a formal impossibility proof, because Backpack sense vectors are input-dependent. It is a geometric warning: practical K values are far smaller than the rank needed to preserve most of the base hidden-state variance.

\begin{table}[t]
  \centering
  \small
  \begin{tabular}{lr}
    \toprule
    Quantity & Value \\
    \midrule
    Tokens analyzed & 10,000 \\
    Hidden dimension & 960 \\
    K=8 cumulative variance & 71.2\% \\
    Canonical-K cumulative variance & 84.9\% \\
    K=64 cumulative variance & 88.3\% \\
    Rank for 95\% variance & 287 \\
    Rank for 99\% variance & 697 \\
    \bottomrule
  \end{tabular}
  \caption{SVD diagnostic for centered SmolLM2-360M final-layer hidden states.}
  \label{tab:svd-diagnostic}
\end{table}

The practical implication is that a converted Backpack with small K must compress structure the base model uses for prediction. The canonical ACROS sense count captures only about 84.9\% of the measured variance, while 95\% variance requires rank 287 and 99\% requires rank 697. This explains why conversion can expose a sense path but still fail as a language model, and motivates a residual design that augments the pretrained output path instead of replacing it.

\section{ACROS Architecture Details}
\label{app:architecture-details}

This appendix expands the implementation details behind the residual architecture in Figure~\ref{fig:architecture}. The reported SmolLM2 ACROS checkpoint wraps the original \texttt{HuggingFaceTB/SmolLM2-360M} decoder without replacing its prediction path: the pretrained backbone and tied LM head remain frozen, and ACROS trains only the residual sense pathway and a scalar gate.

\paragraph{Sense vectors.}
For each token embedding, the sense network produces $K$ vectors in the model hidden dimension. In the SmolLM2 $K{=}32$ run, token embeddings of dimension $d{=}960$ pass through a small residual MLP block, a LayerNorm, and a final MLP whose output dimension is $K d$. The result is reshaped into source-token sense vectors
\begin{equation}
  E \in \mathbf{R}^{B \times K \times T \times d},
\end{equation}
where $B$ is batch size and $T$ is sequence length. These vectors are not vocabulary entries or external dictionary senses; they are token-local residual directions learned from the induction objective.

\paragraph{Contextualization layer.}
The contextualization layer is a lightweight causal query/key attention module, not a full Transformer block. It takes the frozen backbone hidden states $B \in \mathbf{R}^{B \times T \times d}$ and applies one linear map to produce per-sense queries and keys:
\begin{equation}
  Q,K \in \mathbf{R}^{B \times T \times K \times d/K}.
\end{equation}
For each sense slot $k$, it computes causal attention weights over source positions:
\begin{equation}
  C_{k,q,j}
  =
  \mathrm{softmax}_{j \le q}
  \left(
    \frac{Q_{q,k}^{\top}K_{j,k}}{\sqrt{d/K}}
  \right).
\end{equation}
There is no learned value projection in this layer. The values are the sense vectors $E_{k,j}$ produced from the input embeddings. The residual mixture at query position $q$ is therefore:
\begin{equation}
  M_q = \sum_{k=1}^{K}\sum_{j \le q} C_{k,q,j} E_{k,j}.
\end{equation}

\paragraph{Residual gate.}
The current model uses one learned global scalar gate $g$, shared across all layers, positions, tokens, and sense slots:
\begin{equation}
  H_q = B_q + gM_q,
  \qquad
  z_q = H_q W^\top .
\end{equation}
The gate is initialized to $0$, so the initialized ACROS checkpoint exactly matches the base model before induction. The experiments in this paper keep this simple scalar-gate design throughout. Position-wise, token-wise, layer-wise, or sense-wise gates are natural future extensions, but they are not used in the reported results.

\begin{table}[t]
  \centering
  \footnotesize
  \setlength{\tabcolsep}{3pt}
  \begin{tabularx}{\columnwidth}{@{}lX@{}}
    \toprule
    Setting & Value \\
    \midrule
    Base model & SmolLM2-360M \\
    Sense slots & $K=32$ \\
    Hidden size & 960 \\
    Sense MLP scale & 4 \\
    Frozen modules & Backbone and LM head \\
    Trainable modules & Sense network, contextualization, gate \\
    Induction data & FineWeb English \\
    Sequence length & 2048 \\
    GPUs & 4 A100 \\
    Batch / GPU & 4 \\
    Gradient accumulation & 8 \\
    Optimizer & AdamW \\
    Learning rate & $2{\times}10^{-4}$ \\
    Weight decay & 0.1 \\
    Warmup ratio & 0.02 \\
    Max steps & 40,000 \\
    Precision & bfloat16 \\
    $\alpha$ for CLM/KD mixture & 0.5 \\
    KD temperature & 2.0 \\
    $\lambda_{\mathrm{div}}$ & 0.005 \\
    Gate initialization & 0.0 \\
    Gate learning rate & main learning rate \\
    \bottomrule
  \end{tabularx}
  \caption{Main ACROS induction hyperparameters for the SmolLM2-360M $K{=}32$ checkpoint.}
  \label{tab:acros-hparams}
\end{table}

The main SmolLM2 $K{=}32$ induction run therefore sees approximately
$40{,}000 \times 2048 \times 4 \times 4 \times 8 \approx 10.49$B
training tokens, assuming fully packed sequences. No hyperparameter
search was performed; all values in Table~\ref{tab:acros-hparams} are
the fixed recipe used for the reported induction runs.

\begin{table}[t]
  \centering
  \footnotesize
  \setlength{\tabcolsep}{3pt}
  \begin{tabular}{lrr}
    \toprule
    Run & Wall h & A100 GPU-h \\
    \midrule
    SmolLM2-360M $K{=}32$ & 32.5 & 129.9 \\
    SmolLM2-360M $K{=}8$ & 28.3 & 113.3 \\
    SmolLM2-1.7B $K{=}32$ & 81.2 & 325.0 \\
    Pythia-410M $K{=}32$ & 30.4 & 121.8 \\
    OPT-350M $K{=}32$ & 24.4 & 97.5 \\
    SENSIA adapters (4) & 68.5 & 274.0 \\
    \midrule
    Reported-run subtotal & 265.3 & 1061.5 \\
    Frozen BP $K{=}32$ control & 31.8 & 127.1 \\
    \bottomrule
  \end{tabular}
  \caption{Approximate training compute for reported checkpoints, computed from W\&B runtime artifacts multiplied by the allocated four-A100 recipes. Evaluation-only jobs and local qualitative generation are excluded.}
  \label{tab:training-compute}
\end{table}

\begin{table}[t]
  \centering
  \small
  \setlength{\tabcolsep}{3pt}
  \begin{tabular}{lrr}
    \toprule
    Component & Parameters & Status \\
    \midrule
    Base SmolLM2 & 361.8M & frozen \\
    Sense network & 129.1M & trained \\
    Contextualization layer & 1.8M & trained \\
    Global gate & 1 & trained \\
    Auxiliary projection & 0.9M & frozen / unused \\
    \midrule
    ACROS checkpoint total & 493.7M & -- \\
    Active added path & 130.9M & trained \\
    \bottomrule
  \end{tabular}
  \caption{Parameter counts for the SmolLM2-360M $K{=}32$ ACROS checkpoint. The auxiliary projection is present in the implementation for optional reconstruction warmup, but the reported runs set the warmup to zero, so it is not active in training or inference.}
  \label{tab:acros-params}
\end{table}

\paragraph{Latency.}
ACROS adds three operations to each full forward pass: the sense MLP, the contextualization attention, and the residual addition before the frozen LM head. The implementation used here prioritizes faithful access to the full sense contextualization tensor; generation currently disables key-value caching, so the numbers below should be read as full-sequence scoring latency rather than optimized autoregressive decoding latency. In a local CPU-only fp32 benchmark with batch size 1, ACROS adds about $1.3{\times}$--$1.5{\times}$ full-forward latency over the base SmolLM2 model; on CUDA bf16 the overhead drops to $1.03{\times}$--$1.06{\times}$, reflecting the sense path's low arithmetic intensity relative to the frozen backbone on GPU (Table~\ref{tab:acros-latency}).

\begin{table}[t]
  \centering
  \small
  \setlength{\tabcolsep}{4pt}
  \begin{tabular}{llrrr}
    \toprule
    Backend & Seq. len. & Base ms & ACROS ms & Ratio \\
    \midrule
    CPU fp32 & 64  & 66.7  & 102.2 & 1.53$\times$ \\
    CPU fp32 & 128 & 95.0  & 123.7 & 1.30$\times$ \\
    CPU fp32 & 256 & 157.7 & 201.5 & 1.28$\times$ \\
    \addlinespace
    CUDA bf16 & 64  & 23.8 & 24.7 & 1.04$\times$ \\
    CUDA bf16 & 128 & 23.1 & 23.8 & 1.03$\times$ \\
    CUDA bf16 & 256 & 22.7 & 24.0 & 1.06$\times$ \\
    \bottomrule
  \end{tabular}
  \caption{Full-forward latency for SmolLM2-360M and ACROS $K{=}32$ at batch size 1. CPU fp32 timings are means over repeated forward passes on a local machine. CUDA bf16 timings use an NVIDIA RTX A5000 with 20 warmup and 100 timed forwards using CUDA event timing. Neither backend uses KV-cache; numbers reflect full-sequence scoring latency, not optimized autoregressive decoding.}
  \label{tab:acros-latency}
\end{table}

\section{SENSIA Protocol Details}
\label{app:sensia}

\begin{table}[t]
  \centering
  \small
  \setlength{\tabcolsep}{0pt}
  \begin{tabular*}{\columnwidth}{@{}l@{\extracolsep{\fill}}rrr@{}}
    \toprule
    Model & ROUGE-L & chrF++ & Copy-4g \\
    \midrule
    Base SmolLM2 & 0.088 & 15.30 & 0.368 \\
    ACROS & 0.110 & 18.30 & 0.125 \\
    Frozen BP & 0.013 & 2.81 & 0.000 \\
    Gemma 270M & 0.139 & 25.73 & 0.784 \\
    Qwen2 0.5B & 0.109 & 20.23 & 0.411 \\
    \bottomrule
  \end{tabular*}
  \caption{XL-Sum full-test generation over available languages. ACROS improves over the pure English Base SmolLM2 starting point and Frozen BP, while copying much less than modern dense baselines.}
  \label{tab:xlsum}
\end{table}

\begin{table*}[t]
  \centering
  \scriptsize
  \setlength{\tabcolsep}{2pt}
  \begin{tabular*}{\textwidth}{@{}ll@{\extracolsep{\fill}}ccccc@{}}
    \toprule
    Task & Lang & Base & Frozen BP & ACROS & Gemma & Qwen \\
    \midrule
    XCOPA & ind & .520\tiny{ $\pm$.044} & .514\tiny{ $\pm$.044} & .594\tiny{ $\pm$.043} & .580\tiny{ $\pm$.043} & .558\tiny{ $\pm$.044} \\
    XCOPA & est & .488\tiny{ $\pm$.044} & .518\tiny{ $\pm$.044} & .546\tiny{ $\pm$.044} & .496\tiny{ $\pm$.044} & .486\tiny{ $\pm$.044} \\
    XCOPA & swh & .534\tiny{ $\pm$.044} & .558\tiny{ $\pm$.044} & .550\tiny{ $\pm$.044} & .544\tiny{ $\pm$.044} & .544\tiny{ $\pm$.044} \\
    XCOPA & tur & .538\tiny{ $\pm$.044} & .530\tiny{ $\pm$.044} & .540\tiny{ $\pm$.044} & .570\tiny{ $\pm$.043} & .538\tiny{ $\pm$.044} \\
    \midrule
    Belebele & ind & .281\tiny{ $\pm$.029} & .230\tiny{ $\pm$.028} & .223\tiny{ $\pm$.027} & .259\tiny{ $\pm$.029} & .324\tiny{ $\pm$.031} \\
    Belebele & est & .220\tiny{ $\pm$.027} & .256\tiny{ $\pm$.029} & .229\tiny{ $\pm$.027} & .259\tiny{ $\pm$.029} & .287\tiny{ $\pm$.030} \\
    Belebele & swh & .220\tiny{ $\pm$.027} & .271\tiny{ $\pm$.029} & .237\tiny{ $\pm$.028} & .229\tiny{ $\pm$.027} & .274\tiny{ $\pm$.029} \\
    Belebele & tur & .236\tiny{ $\pm$.028} & .209\tiny{ $\pm$.027} & .234\tiny{ $\pm$.028} & .272\tiny{ $\pm$.029} & .309\tiny{ $\pm$.030} \\
    \midrule
    XStory & ind & .496\tiny{ $\pm$.025} & .484\tiny{ $\pm$.025} & .554\tiny{ $\pm$.025} & .547\tiny{ $\pm$.025} & .543\tiny{ $\pm$.025} \\
    XStory & swh & .488\tiny{ $\pm$.025} & .485\tiny{ $\pm$.025} & .529\tiny{ $\pm$.025} & .507\tiny{ $\pm$.025} & .494\tiny{ $\pm$.025} \\
    \bottomrule
  \end{tabular*}
  \caption{Downstream full-dataset results with approximate 95\% intervals. Intervals are unpaired standard-error estimates for each task--language cell, included to show uncertainty and overlap across ACROS and baseline models.}
  \label{tab:downstream-cis}
\end{table*}

The SENSIA adaptation experiment follows the same intrinsic alignment logic as the original SENSIA setup \citep{cruz-etal-2026-multilinguality}. We adapt from English into Indonesian, Estonian, Swahili, and Turkish, then average contextual retrieval, sense retrieval, and target-language FLORES PPL across languages.

\paragraph{Adaptation recipe.}
SENSIA adaptation starts from the English ACROS checkpoint and fine-tunes on English--target parallel data using a three-phase scheduled objective combining context InfoNCE, sense InfoNCE, and target LM losses with label smoothing ($\epsilon{=}0.05$). The alignment phase (first 20\% of steps) emphasizes retrieval objectives; the middle phase (30\%) balances retrieval and LM; the polish phase (final 50\%) emphasizes target LM while freezing the sense network and sense weight net. All runs use AdamW with learning rate $5{\times}10^{-5}$, linear warmup over 2\% of steps, global batch size 64 (16 per GPU, 4${\times}$A100), sequence length 256, and 150,000 total steps in bfloat16 mixed precision. Frozen BP uses the same data, objective schedule, step count, and batch recipe, but starts from legacy Backpack checkpoints and differs in parameter trainability and precision defaults; it is included as a degraded-substrate control, not as an identical ACROS variant.

\paragraph{Parallel data.}
We use the same English--target parallel pairs filtered from OPUS and made available by \citet{cruz-etal-2026-multilinguality}. Pairs are filtered by length, language identification, LaBSE similarity, and deduplication. Corpora drawn on include CCMatrix, CCAligned, ParaCrawl, JW300, OpenSubtitles, WikiMatrix, TED2020, GlobalVoices, and language-specific collections including EUbookshop and DGT for Estonian, SETimes for Turkish, and JW300 and Tanzil for Swahili.

\paragraph{Intrinsic retrieval and confidence intervals.}
Target-language PPL and retrieval both use the FLORES-200 \texttt{devtest} split (1,012 examples per language). Retrieval is bidirectional cosine nearest-neighbor R@1 with no re-ranking: source-to-target and target-to-source R@1 are averaged. Context embeddings are the last non-pad hidden state; sense embeddings use SENSIA sense pooling at temperature 0.7. Bootstrap CIs use 10,000 resamples (seed 20260514) over sentence pairs, language-balanced across the four target languages.

\paragraph{XL-Sum generation.}
Summaries are generated with greedy decoding, \texttt{max\_input\_tokens=512}, and \texttt{max\_new\_tokens=80}, using language-specific completion prompts (no chat template). Estonian is unavailable in XL-Sum; results cover Indonesian (4,780 test examples), Swahili (987), and Turkish (3,397). ACROS and Frozen BP decode without KV-cache due to implementation constraints; dense baselines use KV-cache when supported. ROUGE-L is \texttt{rouge-score~0.1.2} F1; chrF++ is \texttt{sacrebleu~2.6.0} with word order 2. Copy-4g is 4-gram precision of the generation against the source article; Rep-3g is the repeated 3-gram rate; Empty is the fraction of empty generations.

\paragraph{Downstream likelihood evaluation.}
XCOPA, XStoryCloze, and Belebele are evaluated zero-shot with the Eleuther LM Evaluation Harness in likelihood-ranking mode (not generation). XCOPA and XStoryCloze have 2 choices (50\% chance); Belebele has 4 choices (25\% chance). XStoryCloze covers Indonesian and Swahili only; Estonian and Turkish are unavailable. Per-cell intervals in Table~\ref{tab:downstream-cis} are $\mathrm{acc}\pm1.96{\times}\mathrm{stderr}$, unpaired.

\paragraph{Baselines.}
Gemma 3 270M uses \texttt{google/gemma-3-270m} (base, not instruct) in bfloat16 with completion prompts. Qwen2 0.5B uses \texttt{Qwen/Qwen2-0.5B} (base, not instruct) with completion prompts. Both use KV-cache for generation.

\section{Downstream Standard-Error Intervals}
\label{app:downstream-cis}

Table~\ref{tab:downstream-cis} gives approximate 95\% intervals ($\mathrm{acc}\pm 1.96\times\mathrm{stderr}$) from the Eleuther evaluation harness for each task--language cell. These are not paired bootstrap tests and should not be used for strong significance claims; they are included so that readers can judge the overlap pattern referenced in the main text.

\section{Further Architectural Ablations}
\label{app:capacity-scale}

The following ablations characterize boundary conditions of the ACROS induction recipe along two dimensions: sense-slot capacity and backbone scale. They are not central to the main claims of the paper. All checkpoints are single runs under the fixed induction recipe; repeated-trial variance estimates are not reported for these rows.

\begin{table}[t]
  \centering
  {\small
  \begin{tabular}{lrr}
    \toprule
    Metric & $K=8$ & $K=32$ \\
    \midrule
    WSD F1 & 64.91 & 64.95 \\
    CoInCo delta & \textbf{+0.002018} & +0.001642 \\
    CoInCo KL & 1.13e-6 & \textbf{5.69e-7} \\
    \bottomrule
  \end{tabular}}
  \caption{K capacity ablation. Current probes do not establish a monotonic ``more senses is better'' trend.}
  \label{tab:k}
\end{table}

\begin{table}[t]
  \centering
  {\small
  \setlength{\tabcolsep}{0pt}
  \begin{tabular*}{\columnwidth}{@{}l@{\extracolsep{\fill}}rr@{}}
    \toprule
    Metric & 360M & 1.7B \\
    \midrule
    FLORES PPL & 25.09 & \textbf{20.41} \\
    LAMA @1/@5 & 0.3175/0.5245 & \textbf{0.3550/0.5770} \\
    Gate value & -0.05090 & -0.01813 \\
    Contrib. ratio & 1.745\% & 1.548\% \\
    WS-353 mean/max & \textbf{0.449/0.487} & 0.270/0.328 \\
    WSD F1 & \textbf{64.95} & 64.18 \\
    CoInCo delta & \textbf{+0.001642} & +0.000893 \\
    \bottomrule
  \end{tabular*}}
  \caption{Scale and diversity ablation. The 1.7B model improves ordinary LM health but not the current sense-interface metrics; a tenfold increase in $\lambda_{\mathrm{div}}$ raises slot orthogonality but does not recover WS-353 or WSD, ruling out sense collapse as the bottleneck.}
  \label{tab:scale}
\end{table}

\subsection{Sense-Slot Capacity}

\paragraph{Research Question.} Does increasing the number of sense slots $K$ monotonically improve sense-interface quality under fixed induction?

\paragraph{Methodology.} We train matched single-run 360M ACROS checkpoints with $K=8$ and $K=32$ and evaluate with the same WSD and CoInCo protocols (Table~\ref{tab:k}).

\paragraph{Results.} WSD is effectively tied across both settings (64.91 vs.\ 64.95 F1). CoInCo shows a tradeoff: $K{=}8$ yields larger substitute-mass delta while $K{=}32$ yields lower KL. Increasing $K$ changes the granularity of the interface rather than monotonically improving it under the current probes.

\subsection{Backbone Scale and Induction Pressure}

\paragraph{Research Question.} Does the residual ACROS recipe improve when the backbone is scaled from SmolLM2-360M to SmolLM2-1.7B? And does increasing diversity pressure explain any observed degradation?

\paragraph{Methodology.} We compare single-run 360M and 1.7B $K{=}32$ ACROS checkpoints on LM health (English FLORES PPL, LAMA) and sense-interface quality (gate, contribution ratio, WS-353, WSD, and CoInCo target-best steering). To diagnose the source of any degradation, we additionally train a matched 1.7B checkpoint with $\lambda_{\mathrm{div}}$ increased tenfold (0.005$\to$0.05), holding all other hyperparameters fixed.

\paragraph{Results.} The 1.7B backbone is a better language model (FLORES PPL drops from 25.09 to 20.41; LAMA top-1/top-5 rise to 0.3550/0.5770), but the sense interface does not proportionally improve: WS-353, WSD, and CoInCo all weaken relative to 360M. To diagnose the cause, we ran a matched 1.7B checkpoint with $\lambda_{\mathrm{div}}$ increased tenfold (0.005$\to$0.05). Slot orthogonality and contribution ratio both increase under stronger diversity pressure, confirming that the sense vectors are not collapsing. However, WS-353 max and WSD do not recover, which rules out sense collapse as the primary bottleneck. The limiting factor appears to be induction signal strength: a more capable backbone satisfies the objective with less residual contribution, leaving the gate less open regardless of slot structure. This suggests a concrete design direction: objectives that specifically reward gate activation, such as sense-reconstruction losses or gate-magnitude terms, are likely to extend the current recipe to larger backbones. Systematic exploration of such objectives is out of scope for the current paper and is left for future work.

\section{Prompt Continuation Samples}
\label{app:prompt-continuations}

Tables~\ref{tab:prompt-eng}--\ref{tab:prompt-downstream} show illustrative continuations from the English ACROS checkpoint and the four SENSIA-adapted checkpoints. Prompts are translated or lightly adapted across languages. Outputs were generated with greedy decoding and top-$p$ sampling, then lightly trimmed to remove terminal fragments and repeated punctuation.

\begin{table*}[!t]
  \centering
  \scriptsize
  \setlength{\tabcolsep}{3pt}
  \renewcommand{\arraystretch}{1.08}
  \begin{tabularx}{\textwidth}{@{}>{\raggedright\arraybackslash}p{0.23\textwidth}>{\raggedright\arraybackslash}X>{\raggedright\arraybackslash}X@{}}
    \toprule
    Prompt & Greedy & Sampled \\
    \midrule
    The president of the country announced that &
    The president of the country announced that the country will be free from the debt crisis. &
    The president of the country announced that there will be a referendum on the constitution. If it does not happen by March 19, the country will hold a national election on April 19. \\
    The hospital reported that the number of patients &
    The hospital reported that the number of patients with the disease has increased by 100\% in the past year. &
    The hospital reported that the number of patients with hepatitis C has decreased by 60\% in the past two years due to the drug, but 20\% of the people who receive hepatitis C treatments develop drug resistance. \\
    Researchers at the university discovered a new &
    Researchers at the university discovered a new way to make a material that can be used to make a variety of different devices, including a new type of solar cell. &
    Researchers at the university discovered a new technique that can convert water to hydrogen gas. This breakthrough, if perfected, will make it much easier to store and transport hydrogen fuel. \\
    \bottomrule
  \end{tabularx}
  \caption{English ACROS continuation samples.}
  \label{tab:prompt-eng}
\end{table*}

\begin{table*}[!t]
  \centering
  \scriptsize
  \setlength{\tabcolsep}{2pt}
  \renewcommand{\arraystretch}{1.04}
  \begin{tabularx}{\textwidth}{@{}>{\raggedright\arraybackslash}p{0.045\textwidth}>{\raggedright\arraybackslash}p{0.19\textwidth}>{\raggedright\arraybackslash}X>{\raggedright\arraybackslash}X@{}}
    \toprule
    Lang. & Prompt & Greedy & Sampled \\
    \midrule
    est & Riigi president teatas, et\par\vspace{1pt}\emph{(EN: The president of the country announced that)} &
    Riigi president teatas, et ta ei ole valmis presidendiks saamiseks.\par\vspace{1pt}\emph{(EN: The president of the country announced that they are not ready to become president.)} &
    Riigi president teatas, et tahab 2020. aastal kinnitada seadusi, mis reguleerivad poliitikat.\par\vspace{1pt}\emph{(EN: The president of the country announced that they want to approve laws in 2020 that regulate policy.)} \\
    est & Haigla teatas, et patsientide arv\par\vspace{1pt}\emph{(EN: The hospital reported that the number of patients)} &
    Haigla teatas, et patsientide arv on 1000.\par\vspace{1pt}\emph{(EN: The hospital reported that the number of patients is 1000.)} &
    Haigla teatas, et patsientide arv oli umbes 3000, kellest enamus olid pärit Euroopa Liidu osariikidest.\par\vspace{1pt}\emph{(EN: The hospital reported that the number of patients was about 3000, most from European Union member states.)} \\
    est & Ülikooli teadlased avastasid uue\par\vspace{1pt}\emph{(EN: Researchers at the university discovered a new)} &
    Ülikooli teadlased avastasid uue koloonia, mis on kõige suurem ja kõige väiksem kogu maailmas.\par\vspace{1pt}\emph{(EN: Researchers at the university discovered a new colony that is the largest and smallest in the world.)} &
    Ülikooli teadlased avastasid uue tüüpi muutujad, mis mõjutavad looduslikku soojuskiirgust.\par\vspace{1pt}\emph{(EN: Researchers at the university discovered new types of variables that affect natural thermal radiation.)} \\
    \addlinespace[2pt]
    \midrule
    ind & Presiden negara itu mengumumkan bahwa\par\vspace{1pt}\emph{(EN: The president of the country announced that)} &
    Presiden negara itu mengumumkan bahwa ia akan menjadi presiden pertama yang menjabat dari 2019 hingga 2024.\par\vspace{1pt}\emph{(EN: The president of the country announced that he would be the first president to serve from 2019 to 2024.)} &
    Presiden negara itu mengumumkan bahwa keputusan itu telah dirumuskan oleh Komite Eropa untuk Kepentingan Urusan Timur Tengah.\par\vspace{1pt}\emph{(EN: The president announced that the decision had been formulated by the European Committee for Middle Eastern Affairs.)} \\
    ind & Rumah sakit melaporkan bahwa jumlah pasien\par\vspace{1pt}\emph{(EN: The hospital reported that the number of patients)} &
    Rumah sakit melaporkan bahwa jumlah pasien yang terinfeksi meningkat dari 100 orang pada 2015 menjadi 1.000 orang.\par\vspace{1pt}\emph{(EN: The hospital reported that infected patients increased from 100 people in 2015 to 1,000.)} &
    Rumah sakit melaporkan bahwa jumlah pasien meningkat 15\% selama periode 10 hari sebelum dan sesudah penyelidikan.\par\vspace{1pt}\emph{(EN: The hospital reported that the number of patients increased by 15\% during the 10-day period before and after the investigation.)} \\
    ind & Para peneliti di universitas menemukan sebuah\par\vspace{1pt}\emph{(EN: Researchers at the university discovered a)} &
    Para peneliti di universitas menemukan sebuah komunitas di mana 100\% dari mereka yang terinfeksi virus itu mengalami komplikasi.\par\vspace{1pt}\emph{(EN: Researchers found a community where 100\% of those infected with the virus experienced complications.)} &
    Para peneliti di universitas menemukan sebuah sistem kekebalan tubuh yang kuat yang dapat melawan virus, bakteri, dan parasit.\par\vspace{1pt}\emph{(EN: Researchers discovered a strong immune system that can fight viruses, bacteria, and parasites.)} \\
    \addlinespace[2pt]
    \midrule
    swh & Rais wa nchi alitangaza kwamba\par\vspace{1pt}\emph{(EN: The president of the country announced that)} &
    Rais wa nchi alitangaza kwamba, katika kipindi cha miezi 12 ijayo, nchi itaendelea kuwa na uchumi wa kutosha.\par\vspace{1pt}\emph{(EN: The president announced that, over the next 12 months, the country would continue to have a sufficient economy.)} &
    Rais wa nchi alitangaza kwamba amewafukuza viongozi wa makundi yenye silaha wakiwa katika kambi za jeshi.\par\vspace{1pt}\emph{(EN: The president announced that he had expelled leaders of armed groups while they were in army camps.)} \\
    swh & Hospitali iliripoti kwamba idadi ya wagonjwa\par\vspace{1pt}\emph{(EN: The hospital reported that the number of patients)} &
    Hospitali iliripoti kwamba idadi ya wagonjwa wa Ebola nchini DRC imepungua kwa asilimia 90 tangu mwezi Aprili mwaka huu.\par\vspace{1pt}\emph{(EN: The hospital reported that the number of Ebola patients in the DRC has decreased by 90 percent since April this year.)} &
    Hospitali iliripoti kwamba idadi ya wagonjwa wanaopata ujauzito ni sawa na ilivyokuwa wakati wa mwaka 2016.\par\vspace{1pt}\emph{(EN: The hospital reported that the number of patients becoming pregnant is the same as it was in 2016.)} \\
    swh & Watafiti katika chuo kikuu waligundua aina mpya ya\par\vspace{1pt}\emph{(EN: Researchers at the university discovered a new type of)} &
    Watafiti katika chuo kikuu waligundua aina mpya ya virusi vya corona, ambavyo vilikuwa vimeenea kwa kasi katika miji mingi.\par\vspace{1pt}\emph{(EN: Researchers discovered a new type of coronavirus that had spread rapidly in many cities.)} &
    Watafiti katika chuo kikuu waligundua aina mpya ya virusi vya homa ya nguruwe inayoitwa canine parainfluenza.\par\vspace{1pt}\emph{(EN: Researchers discovered a new type of swine flu virus called canine parainfluenza.)} \\
    \addlinespace[2pt]
    \midrule
    tur & Ülkenin cumhurbaşkanı şunu duyurdu:\par\vspace{1pt}\emph{(EN: The president of the country announced:)} &
    Ülkenin cumhurbaşkanı şunu duyurdu: ``Bizim için, ülkemizin kültürü, ekonomisi ve toplumu\ldots''\par\vspace{1pt}\emph{(EN: The country's president announced: ``For us, our country's culture, economy, and society\ldots'')} &
    Ülkenin cumhurbaşkanı şunu duyurdu: ``Yeni bir anayasa kabul edildi, eğer iktidarı yeniden kurmaya çalışırsak, 2013'te başlayacağız.''\par\vspace{1pt}\emph{(EN: The country's president announced: ``A new constitution has been adopted; if we try to reestablish power, we will start in 2013.'')} \\
    tur & Hastane, hasta sayısının\par\vspace{1pt}\emph{(EN: The hospital, the number of patients)} &
    Hastane, hasta sayısının 1000'e ulaştığını gösterdi. Ve hastanede 1000 kişi var.\par\vspace{1pt}\emph{(EN: The hospital showed that the number of patients reached 1000. And there are 1000 people in the hospital.)} &
    Hastane, hasta sayısının sürekli artmasıyla, 12 hasta tespit edildi.\par\vspace{1pt}\emph{(EN: As the hospital's number of patients continued to increase, 12 patients were detected.)} \\
    tur & Üniversitedeki araştırmacılar yeni bir keşif yaptı:\par\vspace{1pt}\emph{(EN: Researchers at the university made a new discovery:)} &
    Üniversitedeki araştırmacılar yeni bir keşif yaptı: Dünya'nın yüzeyinde bulunan 100 milyon yıllık bir kaya oluşumu.\par\vspace{1pt}\emph{(EN: Researchers made a new discovery: a 100-million-year-old rock formation found on Earth's surface.)} &
    Üniversitedeki araştırmacılar yeni bir keşif yaptı: Bu nesnelerin insan yapımı olduğunu keşfettiler.\par\vspace{1pt}\emph{(EN: Researchers made a new discovery: they discovered that these objects were human-made.)} \\
    \bottomrule
  \end{tabularx}
  \caption{SENSIA-adapted ACROS continuation samples for the four downstream languages.}
  \label{tab:prompt-downstream}
\end{table*}

\section{XL-Sum Qualitative Summaries}
\label{app:xlsum-qualitative}

\begin{table*}[!t]
  \centering
  \scriptsize
  \setlength{\tabcolsep}{2pt}
  \renewcommand{\arraystretch}{0.90}
  \begin{tabularx}{\textwidth}{@{}>{\raggedright\arraybackslash}p{0.11\textwidth}>{\raggedright\arraybackslash}p{0.25\textwidth}>{\raggedright\arraybackslash}X>{\raggedright\arraybackslash}p{0.12\textwidth}@{}}
    \toprule
    Example & Reference & Model output & Metrics \\
    \midrule
    Indonesian: Kuala Lumpur grenade blast &
    Satu orang tewas dan 12 orang lainnya terluka dalam ledakan di distrik hiburan populer di ibukota Malaysia, Kuala Lumpur.\par\vspace{1pt}\emph{(EN: One person was killed and 12 others were injured in an explosion in a popular entertainment district in Malaysia's capital, Kuala Lumpur.)} &
    \textbf{ACROS:} Penyelidikan polisi menunjukkan bahwa insiden terjadi di kawasan Bukit Bintang, Malaysia. Selain itu, polisi juga menemukan bahan peledak di mobil yang sedang diparkir di luar sebuah toko di luar kawasan Bukit Bintang.\par\vspace{1pt}\emph{(EN: Police say the incident occurred in Bukit Bintang and that explosives were found in a parked car nearby.)} &
    R-L=.078; chrF=22.3; Copy=.069; Rep=.033 \\
    &
    &
    \textbf{Gemma:} Polisi Malaysia melaporkan bahwa sebuah granat diduga dilemparkan ke jalan di luar bar di kawasan Bukit Bintang pada Kamis pagi. Satu bahan peledak lain ditemukan dibawah sebuah mobil yang sedang diparkir tetapi tidak meledak.\par\vspace{1pt}\emph{(EN: Malaysian police report that a grenade was thrown outside a bar and another explosive was found under a parked car.)} &
    R-L=.108; chrF=21.6; Copy=.904; Rep=.000 \\
    &
    &
    \textbf{Qwen:} Ledakan terjadi pada Kamis pagi di Bukit Bintang, Malaysia; granat diduga dilemparkan ke jalan di luar bar, dan satu bahan peledak lain ditemukan di mobil yang sedang diparkir.\par\vspace{1pt}\emph{(EN: An explosion occurred in Bukit Bintang; a grenade was thrown outside a bar and another explosive was found in a parked car.)} &
    R-L=.143; chrF=24.6; Copy=.559; Rep=.057 \\
    \addlinespace[1pt]
    \midrule
    Indonesian: Cambridge written tests &
    Para calon mahasiswa Universitas Cambridge, Inggris, diwajibkan mengikuti ujian tertulis, yang merupakan komponen baru dari proses seleksi, kata direktur pendaftaran perguruan tinggi tersebut, Dr Sam Lucy.\par\vspace{1pt}\emph{(EN: Cambridge applicants in England must take a written exam as a new component of admissions, said admissions director Dr Sam Lucy.)} &
    \textbf{ACROS:} Universitas Cambridge mengatakan tes tertulis akan dilaksanakan bersamaan dengan wawancara atau sebelumnya. Lucy mengatakan ujian tertulis akan disesuaikan dengan setiap jurusan dan akan dilaksanakan pada November tahun ini.\par\vspace{1pt}\emph{(EN: Cambridge says written tests will be held with or before interviews and tailored to each subject.)} &
    R-L=.148; chrF=28.5; Copy=.320; Rep=.000 \\
    &
    &
    \textbf{Gemma:} Universitas Cambridge mengatakan tes tertulis dilaksanakan bersamaan dengan wawancara atau sebelumnya. Rincian tentang ujian tertulis ini dijelaskan Dr Lucy dalam surat yang ia kirim kepada semua sekolah menengah.\par\vspace{1pt}\emph{(EN: Cambridge says written tests will be held with or before interviews, with details sent by Dr Lucy to secondary schools.)} &
    R-L=.171; chrF=34.6; Copy=.981; Rep=.000 \\
    &
    &
    \textbf{Qwen:} Universitas Cambridge mengatakan tes tertulis dilaksanakan bersamaan dengan wawancara atau sebelumnya. Ujian tertulis akan disesuaikan untuk tiap jurusan dan akan dilaksanakan ketika calon mahasiswa mengikuti wawancara.\par\vspace{1pt}\emph{(EN: Cambridge says written tests will be customized by subject and held when applicants attend interviews or before them.)} &
    R-L=.197; chrF=35.2; Copy=.875; Rep=.000 \\
    \addlinespace[1pt]
    \midrule
    Swahili: Somalia and al-Shabab &
    Rais wa Somalia Mohamed Abdullahi "Farmajo" ametangaza vita dhidi ya kundi la wanamgambo la al-Shabab.\par\vspace{1pt}\emph{(EN: Somali President Mohamed Abdullahi "Farmajo" declared war against the al-Shabab militant group.)} &
    \textbf{ACROS:} Rais wa Somalia Mohamed Abdullahi "Farmajo" ametangaza vita dhidi ya kundi la al-Shabab. Ametoa msamaha wa siku 60 kwa wanamgambo hao, akiwaapisha kwa kuwa ni msamaha wa kimataifa. Rais ametangaza pia\par\vspace{1pt}\emph{(EN: Farmajo declared war against al-Shabab and gave militants a 60-day amnesty; the output then trails off.)} &
    R-L=.583; chrF=78.2; Copy=.448; Rep=.000 \\
    &
    &
    \textbf{Gemma:} Rais wa Somalia Mohamed Abdullahi "Farmajo" ametangaza vita dhidi ya kundi wa wanamgambo la al-Shabab. Ametoa pia msamaha wa siku 60 kwa wanamgambo hao, akiwataka kujisalimisha, ili wapate mafunzo, ajira na elimu. Hi ni baada ya shambulizi la\par\vspace{1pt}\emph{(EN: Farmajo declared war against al-Shabab, then closely follows the article's amnesty and surrender details before cutting off.)} &
    R-L=.545; chrF=74.6; Copy=1.000; Rep=.000 \\
    &
    &
    \textbf{Qwen:} "Makala: Rais wa Somalia Mohamed Abdullahi "Farmajo" ametangaza vita dhidi ya kundi wa wanamgambo la al-Shabab. Ametoa pia msamaha wa siku 60 kwa wanamgambo hao, akiwataka kujisalimisha, ili wapate mafunzo, ajira\par\vspace{1pt}\emph{(EN: The output echoes the article prompt, then repeats the article's opening al-Shabab and amnesty details.)} &
    R-L=.625; chrF=77.6; Copy=.966; Rep=.000 \\
    \addlinespace[1pt]
    \midrule
    Turkish: Syria Geneva talks &
    ABD Dışişleri Bakanı John Kerry, İsviçre'nin Cenevre kentinde 22 Ocak'ta yapılacak barış görüşmelerinin Suriye'de bir geçiş hükümeti kurulması için "en iyi fırsat" olduğunu söyledi.\par\vspace{1pt}\emph{(EN: U.S. Secretary of State John Kerry said the January 22 Geneva peace talks were the best opportunity to establish a transitional government in Syria.)} &
    \textbf{ACROS:} Suriye krizinin Çözüm yolu engillerle dolu olduğunu söylüyor, ABD Dışişleri Bakanı Kerry, Cenevre görüşmelerinin Suriye krizinin Çözümüne gözlerini aç\par\vspace{1pt}\emph{(EN: Kerry says the path toward resolving the Syria crisis is full of obstacles and connects the Geneva talks to a possible solution.)} &
    R-L=.310; chrF=35.2; Copy=.375; Rep=.000 \\
    &
    &
    \textbf{Gemma:} Suriye'de yaşanan acılara son vermek için bu fırsatı kaçırmak affedilmez olur. ABD Dışişleri Bakanı, Cenevre görüşmelerinin Suriye krizinin çözümüne yardımcı olabileceğini söyledi. ABD Dışişleri Bakanı, Cenevre görüşmelerinin Suriye krizinin çözüm\par\vspace{1pt}\emph{(EN: The output says the chance to end Syria's suffering should not be missed, then repeats that Geneva could help solve the crisis.)} &
    R-L=.250; chrF=31.7; Copy=.455; Rep=.206 \\
    &
    &
    \textbf{Qwen:} Birleşmiş Milletler, Suriye hükümeti ile muhalefetin ilk kez birlikte katılacağını duyurmasından hemen sonra yaptı. Birleşmiş Milletler, ABD ve Rusya, iki tarafın soruna siyasi bir çözüm bulması için aylardır çaba harcıyor. Fakat, İsvi\par\vspace{1pt}\emph{(EN: The output copies the setup about the U.N., Syria's government and opposition, and U.S.-Russian diplomatic efforts before cutting off.)} &
    R-L=.140; chrF=23.9; Copy=.897; Rep=.000 \\
    \bottomrule
  \end{tabularx}
  \caption{Illustrative XL-Sum full-test examples. References are the original XL-Sum summaries; italic \emph{EN} lines are author translations for readability. Output excerpts are lightly shortened and normalized for display. Metrics are ROUGE-L F1, chrF++, source-copy 4-gram precision, and repeated 3-gram rate for each full generated summary.}
  \label{tab:xlsum-qualitative}
\end{table*}

Table~\ref{tab:xlsum-qualitative} shows XL-Sum examples where ACROS copies less source text than Gemma and Qwen. The examples are illustrative; the aggregate copy-rate comparison is reported in Table~\ref{tab:sensia-downstream}.

\section{Artifact Provenance and Licensing}
\label{app:artifact-provenance}

Table~\ref{tab:artifact-provenance} summarizes the external artifacts used in the experiments. License entries follow the public model cards, dataset cards, or project pages available at the time of writing; for collections such as OPUS, the applicable terms vary by component corpus. We use these resources for research evaluation and adaptation only, and we do not redistribute the raw corpora with this paper.

\paragraph{Software.}
XL-Sum generation metrics use \texttt{rouge-score} 0.1.2 for ROUGE-L and \texttt{sacrebleu} 2.6.0 with chrF word order 2 for chrF++. XCOPA, XStoryCloze, and Belebele likelihood evaluations use the Eleuther LM Evaluation Harness 0.4.13.dev0. Bootstrap aggregation and summary statistics use NumPy 2.4.4 and SciPy 1.17.1; the McNemar exact test used for WSD comparisons is implemented directly in the local evaluator. Local reproduction used PyTorch 2.11.0, Transformers 5.5.3, and Datasets 2.21.0.

\paragraph{Data sensitivity.}
Several artifacts contain web or news text and may include naturally occurring names, public events, and demographic references. We did not collect new human-subject data, attempt re-identification, or augment the benchmark corpora with private information. All artifacts were used for research purposes only, consistent with their access conditions, and no derived artifacts are used outside research contexts.

\begin{table*}[!t]
  \centering
  \scriptsize
  \setlength{\tabcolsep}{3pt}
  \renewcommand{\arraystretch}{1.08}
  \begin{tabularx}{\textwidth}{@{}>{\raggedright\arraybackslash}p{0.18\textwidth}>{\raggedright\arraybackslash}p{0.13\textwidth}>{\raggedright\arraybackslash}p{0.21\textwidth}>{\raggedright\arraybackslash}X@{}}
    \toprule
    Artifact & Type & License / access terms & Use in this paper \\
    \midrule
    FineWeb & Web text corpus & ODC-By & English induction data for ACROS; also used for the hidden-state SVD sample. \\
    SmolLM2-360M & Base LM & Apache-2.0 & Main frozen backbone and teacher for ACROS induction; Base SmolLM2 comparison row. \\
    SmolLM2-1.7B & Base LM & Apache-2.0 & Scale ablation backbone. \\
    Pythia-410M & Base LM & Apache-2.0 & Architecture-portability backbone. \\
    OPT-350M & Base LM & OPT license / model-card terms & Architecture-portability backbone. \\
    Raganato ALL & WSD benchmark & Component Senseval/SemEval and WordNet access terms & Zero-shot word-sense disambiguation evaluation. \\
    CoInCo & Lexical-substitution corpus & CC-BY-3.0-US & Lexical steering evaluation over filtered public test cases. \\
    FLORES-101/200 & Multilingual sentence benchmark & CC-BY-SA 4.0 & English perplexity preservation; SENSIA retrieval and target-language perplexity. \\
    OPUS parallel data & Parallel corpus collection & Mixed per-corpus licenses & English--target parallel pairs for SENSIA adaptation. \\
    XCOPA & Commonsense benchmark & CC-BY 4.0 & Downstream likelihood evaluation. \\
    XStoryCloze & Commonsense benchmark & CC-BY-SA 4.0 & Downstream likelihood evaluation. \\
    Belebele & Reading-comprehension benchmark & CC-BY-SA 4.0 & Downstream likelihood evaluation. \\
    XL-Sum & Summarization benchmark & CC-BY-NC-SA 4.0, research-only & Full-test generation evaluation and qualitative summary examples. \\
    \bottomrule
  \end{tabularx}
  \caption{Artifact provenance for external datasets and pretrained models. ``License / access terms'' is stated at the artifact level; collection resources may inherit additional restrictions from their component sources.}
  \label{tab:artifact-provenance}
\end{table*}

\end{document}